\documentclass[final]{nesy2026} 

\usepackage{rotating}
\usepackage{longtable}

\usepackage{booktabs}
\usepackage[load-configurations=version-1]{siunitx} 

\theorembodyfont{\upshape}
\theoremheaderfont{\scshape}
\theorempostheader{:}
\theoremsep{\newline}

\title[Sample-efficient Neuro-symbolic Proximal Policy Optimization]{Sample-efficient Neuro-symbolic Proximal Policy Optimization}

\clearauthor{
  \Name{Simone Murari}\thanks{These authors contributed equally} \Email{simone.murari@univr.it} \\
  \Name{Celeste Veronese}\footnotemark[1] \thanks{Corresponding author}\Email{celeste.veronese@univr.it} \\
  \Name{Daniele Meli} \Email{daniele.meli@univr.it}\\
  \addr Dept. of Computer Science, University of Verona, Italy
}

\begin{document}

\maketitle

\begin{abstract}
Deep Reinforcement Learning (DRL) algorithms often require a large amount of data and struggle in sparse-reward domains with long planning horizons and multiple sub-goals.
In this paper, we propose a neuro-symbolic extension of Proximal Policy Optimization (PPO) that transfers partial logical policy specifications learned in easier instances to guide learning in more challenging settings.
We introduce two integrations of symbolic guidance: (i) \textsc{H-PPO-Product}, which biases the action distribution at sampling time, and (ii) \textsc{H-PPO-SymLoss}, which augments the PPO loss with a symbolic regularization term.
We evaluate our methods on three benchmarks (OfficeWorld, WaterWorld, and DoorKey), showing consistently faster learning and higher return at convergence than PPO and a Reward Machine baseline, also under imperfect symbolic knowledge.
\end{abstract}

\section{Introduction}
\label{sec:intro}

Deep Reinforcement Learning (DRL) \citep{SuttonBarto2018} is widely applied to solve sequential decision-making problems. It offers significant benefits for complex domains, ranging from robotics \citep{ibarz2021train} to sustainability \citep{zuccotto2024reinforcement, kaage2025reinforcement}.
Despite these successes, scaling DRL to more challenging configurations remains difficult.
First, neural policies act as black boxes, limiting their interpretability and making it hard to extract formal guarantees \citep{vouros2022explainable}.
Furthermore, DRL suffers from \emph{sample inefficiency} \citep{dulac2021challenges}. Agents often require a huge number of interactions to converge to an optimal policy.
This drawback becomes especially evident in environments featuring long horizons, sparse rewards, and complex sequences of sub-goals.

Standard methods attempt to solve this by applying architectural inductive biases \citep{wang2019generalization} or by injecting domain knowledge through Reward Machines (RMs) \citep{icarte2022reward}. However, reliance on scalar reward shaping is often computationally expensive and can degrade performance if the provided heuristics are inaccurate \citep{cheng2021heuristic}.

We propose a neuro-symbolic approach to improve the sample efficiency and scalability of Proximal Policy Optimization (PPO) algorithms \citep{schulman2017proximal}. Our framework uses logical policy representations as an action-level prior to guide PPO training in harder configurations, where standard methods typically fail.

During training, an online reasoning mechanism translates the agent's raw observations into high-level logical predicates, capturing object states and spatial relationships. These predicates are then evaluated against the previously extracted rules to deduce the most promising actions at each step. We integrate these preferences into PPO via two distinct methods: \textsc{H-PPO-Product}, which biases the learned policy towards the suggested actions, hence modifying the action sampling distribution; and \textsc{H-PPO-SymLoss}, which adds a symbolic penalty during optimization updates. Our approach is designed to work with imperfect logical rules \citep{hazra2023deep} and prevents the need to retune PPO hyperparameters when the domain scale increases.

Our main contributions are the following:
\begin{itemize}
    \item A sample-efficient neuro-symbolic PPO framework that leverages partial logical rules to guide the agent toward logically entailed actions, employing either sampling distribution biases or symbolic optimization penalties.
    \item An empirical validation on three well-established gridworld domains: \emph{OfficeWorld}, \emph{WaterWorld} by \cite{icarte2022reward}, and \emph{DoorKey} by \cite{MinigridMiniworld23}. We show substantial improvements in sampling efficiency and return at convergence over standard baselines and RMs, even in the hardest task instances and without hyperparameter tuning.
\end{itemize}

\section{Related Works}
\label{sec:related_works}

Sample inefficiency consistently restricts the scalability and applicability of algorithms like PPO, particularly in sparse-reward environments with sub-goals and long planning horizons. Traditional approaches to mitigate this issue remain purely neural, often improving data efficiency by explicitly decoupling the training of the policy and the value function to enable more aggressive sample reuse \citep{cobbe2021phasic, raileanu2021decoupling}. However, these frameworks suffer from inherent opacity, lack a natural mechanism to directly integrate human-readable domain rules and require careful tuning of the hyperparameters.

Neuro-symbolic integrations offer a way to combine logical reasoning with neural learning to augment interpretability and generalization. Many methods translate formal task specifications into sequential automata, resulting in policy-correction mechanisms built from trajectories \citep{yang2024probabilistic} or automated reward shaping \citep{de2019foundations,furelos2021induction}. Along this line, plan-based reward shaping and reward machines can still help when plans are incomplete or partly incorrect, using potential-based signals to speed up learning in long-sequence sparse-reward tasks \citep{muller2025using} or guiding exploration \citep{icarte2022reward, furelos2021induction}. Yet reward-based interventions are highly sensitive to poorly-specified heuristics \citep{cheng2021heuristic}, which has pushed recent work toward mining logical rules directly from past trajectories \citep{meli2024learning,sreedharan2023optimistic,veronese2024online,veronese2025learning}.

Our method builds upon the principle of \emph{semantic loss} presented in \cite{xu2018semantic}, which encodes symbolic constraints over a neural network's output as a differentiable penalty. In our setting, we leverage the same general principle but target deep reinforcement learning and policy-gradient optimization. We design a term defined through a symbolic policy, composed of logical rules over predicates, and we study the role of a dedicated weight $\Theta$ to balance symbolic guidance and neural learning during training. Finally, our methods share some similarities with Statistical Relational Learning (SRL) \citep{marra2024statistical} and the neuro-symbolic framework proposed in \cite{hazra2023deep}. However, SRL and pure logical learning methods struggle to scale to complicated domains, as they require an accurate definition of the policy search space. By combining the advantages of symbolic reasoning and DRL, our approach overcomes the scalability bottlenecks of the logic-driven policy learning proposed by \cite{hazra2023deep}. As a result, our neuro-symbolic agent can successfully navigate sparse-reward tasks involving extended sequences of sub-goals, where standard DRL algorithms and purely symbolic methods typically fail. Action-level guidance beyond reward modification is inspired from \cite{veronese2026sample,meli2024learning,sreedharan2023optimistic}, which however focused on model-based RL or $\epsilon$-greedy algorithms. Instead, our method is designed for the most advanced and popular class of PPO algorithms for RL. 

\section{Background}
\label{sec:background}

We work in the standard Markov Decision Process (MDP) setting, defined by the tuple $\langle S, A, T, R, \gamma \rangle$: $S$ is the state space, $A$ is the action set, $T(s,a,s')$ is the transition model, $R(s,a)$ is the immediate reward, and $\gamma\in[0,1]$ is the discount factor.
A (stochastic) policy $\pi(a\mid s)$ specifies how actions are sampled given states.
For a trajectory $\tau=(s_0,a_0,s_1,a_1,\dots)$ generated by $\pi$, we define the (discounted) return from time $t$ as
\begin{equation}
\label{eq:discounted_return}
    G_t \;=\; \sum_{k=0}^{\infty} \gamma^k r_{t+k},
\end{equation}
where $r_t = R(s_t,a_t)$.
The objective is to maximize the expected discounted return from the initial state distribution:
\begin{equation}
\label{eq:expected_discounted_return}
    \mathbb{E}_{\tau\sim\pi}\!\left[G_0\right]
\end{equation}

\subsection{Proximal Policy Optimization}

Proximal Policy Optimization (PPO) \citep{schulman2017proximal} is a policy-gradient algorithm that constrains updates through a clipped surrogate objective, yielding stable improvements in practice. Given a batch of trajectories collected under an older policy $\pi_{\text{old}}$, PPO optimizes:
\begin{equation}
\label{eq:ppo_objective}
    \mathcal{L}^{\text{CLIP}}(\theta) = \mathbb{E}_t \left[ \min\left( \rho_t(\theta) \hat{A}_t, \text{clip}(\rho_t(\theta), 1-\epsilon_{\mathrm{clip}}, 1+\epsilon_{\mathrm{clip}}) \hat{A}_t \right) \right]
\end{equation}
where $\rho_t(\theta) = \frac{\pi_\theta(a_t\mid s_t)}{\pi_{\text{old}}(a_t\mid s_t)}$ is the probability ratio, $\epsilon_{\mathrm{clip}}$ is the clipping coefficient (typically 0.2), and $\hat{A}_t$ is the estimated advantage.

The advantage is computed using Generalized Advantage Estimation (GAE, \citet{schulman2018highdimensionalcontinuouscontrolusing}):
\begin{equation}
\label{eq:gae}
    \hat{A}_t = \sum_{l=0}^{\infty} (\gamma \lambda_{\mathrm{GAE}})^l \delta_{t+l}, \quad \text{where} \quad \delta_t = r_t + \gamma V(s_{t+1}) - V(s_t)
\end{equation}
with $\lambda_{\mathrm{GAE}} \in [0,1]$ controlling the bias-variance trade-off.

The complete PPO loss combines the policy loss with a value function loss $\mathcal{L}^{\text{VF}}$ and a bonus on the policy entropy $\mathcal{H}[\pi_\theta]$:
\begin{equation}
\label{eq:ppo_full}
    \mathcal{L}_{\mathrm{PPO}}(\theta) = -\mathcal{L}^{\text{CLIP}}(\theta) + c_1 \mathcal{L}^{\text{VF}}(\theta) - c_2 \mathcal{H}[\pi_\theta]
\end{equation}
where $c_1, c_2$ are scalar coefficients.

\section{Methodology}
\label{sec:methodology}

Before describing our algorithms, we introduce the notion of \emph{action-level} guidance. In contrast with the state heuristics adopted in classical planning \citep{bonet2001planning}, we consider a logical preference over actions, which is more convenient for integration in RL \citep{veronese2026sample,meli2024learning}.
Specifically, the action-level guidance is provided via a \emph{symbolic policy}, expressed in a fragment of first-order logic (namely, Horn clauses).

\begin{definition}[Symbolic Policy]
Consider two sets of logical predicates $\mathcal{F}, \mathcal{A}$, representing abstractions for environmental (state) features and actions in RL, respectively. A \textbf{symbolic policy} $\pi_{sym} : \mathcal{F} \rightarrow \mathcal{A}$ is a set of logical rules over predicates that map environmental features to action terms.
\end{definition}
$\mathcal{F}, \mathcal{A}$ are defined by the task designer.
Formally, we follow classical works in symbolic and neuro-symbolic AI \citep{coradeschi2013short,veronese2026sample,sreedharan2023optimistic} and assume that there exist functions that map the MDP states and actions into sets of grounded logical predicates: $G_{\mathcal{F}} : S \rightarrow \mathcal{G}(\mathcal{F})$ and $G_{\mathcal{A}} : A \rightarrow \mathcal{G}(\mathcal{A})$. We require $G_\mathcal{A}$ to be surjective, since we need each symbolic action to correspond to at least one MDP action. 

The logical policy encodes rules in the form \(\mathtt{a \leftarrow f_1, \dots, f_n}\), where \(\mathtt{f_i}\) are environmental predicates and \(\mathtt{a}\) represents an action.
For example, consider the \emph{DoorKey} domain, where an agent navigates a grid with walls, keys and doors; it must pick up the correct key and open (\emph{toggle}) the door to reach a goal location.
In this setting \(\pi_{sym}\) might contain logical rules like:
\begin{equation*}
    \mathtt{toggle(X) \leftarrow door(X), locked(X), carryingKey(Z), sameColor(X,Z)}
\end{equation*}

To integrate this symbolic knowledge into the RL loop, we define an indicator function \(I_{\pi_{sym}}: S \times A \to \{0,1\}\), such that:
\begin{equation*}
    I_{\pi_{sym}}(s,a) = 1 \iff \pi_{sym}(G_\mathcal{F}(s)) \models G_\mathcal{A}(a) 
\end{equation*}

We now detail the two proposed neuro-symbolic integrations: \textsc{H-PPO-Product} and \textsc{H-PPO-SymLoss}. Both methods leverage the logical guidance provided by \(I_{\pi_{sym}}(s, a)\), but they integrate this knowledge into PPO training differently.

\subsection{\textsc{H-PPO-Product}}
\label{sec:h_ppo_product}

\textsc{H-PPO-Product} (Algorithm \ref{alg:h_ppo_product}) injects symbolic guidance at the \emph{action sampling} stage. At each time step during training, we reshape the current action distribution by reweighting the neural policy toward actions logically entailed by \(\pi_{sym}\), and then we sample from this distribution.

Specifically, we define the following reweighting term (Lines 3-8):
\begin{equation}
    m_t(a, s) = \begin{cases}
    1 + \lambda \cdot \varepsilon_t & \text{if } I_{\pi_{sym}}(s, a) = 1 \\
    1 & \text{otherwise}
    \end{cases}
\end{equation}
where \(\lambda\in(0,1]\) controls how strongly we trust the symbolic policy, and \(\varepsilon_t\) is a scalar that reduces symbolic influence as training progresses following a linear decay: \(\varepsilon_t = \max(\varepsilon_i - t \cdot \varepsilon_r, \varepsilon_f)\).

The guided sampling policy is then the normalized product (Line 9):
\begin{equation}
    \tilde{\pi}_t(a\mid s) = \frac{\pi_\theta(a\mid s)\, m_t(a,s)}{\sum_{a'} \pi_\theta(a'\mid s)\, m_t(a',s)}
\end{equation}
For \(\varepsilon_t\approx 1\), the policy is noticeably biased toward actions with \(I_{\pi_{sym}}(s,a)=1\); for \(\varepsilon_t\to 0\), we recover the original PPO policy \(\pi_\theta\).
This incentivizes early-stage exploration, crucial for robust generalizable learning \citep{ladosz2022exploration}.
In this way, the agent progressively ignores the symbolic policy, and eventually behaves according to the classical PPO neural policy (mitigating the effect of bad or imperfect suggestions).

\begin{algorithm2e}
\RestyleAlgo{ruled}
\LinesNumbered
\DontPrintSemicolon
\SetKwInOut{Require}{Require}

\caption{\textsc{H-PPO-Product} - Guided Action Sampling}
\label{alg:h_ppo_product}

\Require{State \(s_t\), policy \(\pi_\theta\), timestep \(t\)}
\Require{Logical indicator \(I_{\pi_{sym}}(s_t,a)\), confidence \(\lambda\in[0,1]\)}
\Require{Annealing parameters \((\varepsilon_i,\varepsilon_r,\varepsilon_f)\)}
\KwOut{Action \(a_t\), log-probability \(l_t\), entropy \(\mathcal{H}_t\), value \(V_t\)}

\(\pi_{\theta}(\cdot\mid s_t) \leftarrow \text{PolicyNetwork}(s_t; \theta)\)\;

\textcolor{red}{\(\varepsilon_t \leftarrow \max(\varepsilon_i - t\cdot \varepsilon_r,\varepsilon_f)\)}\;

\textcolor{red}{\For{\(a \in A\)}{
    \textcolor{red}{\(m_t(a,s_t)\leftarrow \mathbf{1}\)}\;
    \If{\(I_{\pi_{sym}}(s_t,a)=1\)}{
        \(m_t(a,s_t)\leftarrow m_t(a,s_t)+\lambda\cdot \varepsilon_t\)\;
    }
}}

\textcolor{red}{\(\tilde{\pi}_t(\cdot\mid s_t)\leftarrow \text{Normalize}\big(\pi_\theta(\cdot\mid s_t),  m_t(\cdot,s_t)\big)\)}\;

\(a_t \sim \tilde{\pi}_t(\cdot\mid s_t)\)\;
\(l_t \leftarrow \log \tilde{\pi}_t(a_t\mid s_t)\)\;
\(\mathcal{H}_t \leftarrow \text{Entropy}\big(\tilde{\pi}_t(\cdot\mid s_t)\big)\)\;
\(V_t \leftarrow V_\phi(s_t)\)\;

\Return \(a_t, l_t, \mathcal{H}_t, V_t\)\;
\end{algorithm2e}


\subsection{\texorpdfstring{\textsc{H-PPO-SymLoss}}{\textsc{H-PPO-SymLoss}}}
\label{sec:h_ppo_symloss}

The \textsc{H-PPO-SymLoss} method (Algorithm \ref{alg:h_ppo_symloss}) injects symbolic guidance at the \emph{optimization} stage, by augmenting the standard PPO loss from Equation \eqref{eq:ppo_objective} with a symbolic term that encourages alignment with a reference policy \(\pi_{ref}\) derived from the symbolic policy \(\pi_{sym}\).

We define probabilistic weights for each action based on the logical indicator using a confidence level \(\eta\in(0,1)\) (Line 5) \footnote{$\eta = 1$ would introduce shielding, hence incomplete and sub-optimal search over the action space.}:
\begin{equation}
w_{sym}(a\mid s)=
\begin{cases}
\eta & \text{if } I_{\pi_{sym}}(s,a)=1\\
1-\eta & \text{otherwise}
\end{cases}
\end{equation}
and then normalize over actions to have a novel policy distribution (Line 6):
\begin{equation}
\pi_{ref}(a\mid s)=\frac{w_{sym}(a\mid s)}{\sum_{a'}w_{sym}(a'\mid s)}.
\end{equation}
In this way, logically entailed actions receive higher probability mass, while all actions keep non-zero probability.
Then we calculate \(
    \rho_t^{sym}(\theta) = \frac{\pi_\theta(a_t\mid s_t)}{\pi_{ref}(a_t\mid s_t)}
    \) at Line 7.
In standard PPO, the reference distribution in the ratio is \(\pi_{\theta_{old}}\); here, it is replaced by \(\pi_{ref}\).
We then define a symbolic auxiliary loss \(\mathcal{L}^{sym}\) which follows the same clipped structure as the PPO surrogate in Eq.~\eqref{eq:ppo_objective} (Line 8):

\begin{equation}
\label{eq:lsym}
    \mathcal{L}^{sym} = \mathbb{E}_t \left[ \min\!\left(\rho_t^{sym}(\theta) \hat{A}_t, \text{clip}(\rho_t^{sym}(\theta), 1-\epsilon_{\mathrm{clip}}, 1+\epsilon_{\mathrm{clip}}) \hat{A}_t \right) \right]
\end{equation}

The ratio \(\rho_t^{sym}\) measures the similarity between the neural and reference policies, \(\hat{A}_t\) weighs the update by estimated utility, and clipping limits large updates in the same way as \(\mathcal{L}^{\text{CLIP}}\).

We finally combine \(\mathcal{L}^{sym}\) with the PPO loss from Equation~\eqref{eq:ppo_full}, via a linear parameter $\Theta$ (Line 9):
\begin{equation}
\label{eq:ppo_symloss}
    \mathcal{L}(\theta) = \mathcal{L}_{\mathrm{PPO}}(\theta) - \Theta \cdot \mathcal{L}^{sym}
\end{equation}


\begin{algorithm2e}
\RestyleAlgo{ruled}
\LinesNumbered
\DontPrintSemicolon
\SetKwInOut{Require}{Require}
\footnotesize
\SetAlgoNlRelativeSize{-1}

\caption{\textsc{H-PPO-SymLoss} - Optimization Step}
\label{alg:h_ppo_symloss}

\Require{Batch of transitions \((s_t, a_t, r_t, s_{t+1})\), timestep \(t\)}
\Require{Current policy \(\pi_{\theta}\), Old policy \(\pi_{\theta_{old}}\)}
\Require{Logical indicator \(I_{\pi_{sym}}(s_t,a_t)\), symbolic weight \(\Theta\) (constant) \textbf{or} annealing parameters \((\Theta_i,\Theta_r,\Theta_f)\)}

\(\hat{A}_t \leftarrow\) ComputeGAE\((r_t, V_\phi(s_t), V_\phi(s_{t+1}))\)\;

\(\rho_t \leftarrow \frac{\pi_\theta(a_t\mid s_t)}{\pi_{\theta_{old}}(a_t\mid s_t)}\)\;
\(\mathcal{L}^{CLIP} \leftarrow \min\left( \text{ratio} \cdot \hat{A}_t,\, \text{clip}(\text{ratio}, 1-\epsilon_{\mathrm{clip}}, 1+\epsilon_{\mathrm{clip}}) \cdot \hat{A}_t \right)\)\;

\textcolor{red}{\(\Theta_t = \max(\Theta_i - t \cdot \Theta_r, \Theta_f) \text{ \textbf{or} } \Theta\)}\;

\textcolor{red}{\(w_{sym}(a_t\mid s_t) \leftarrow \eta\) \textbf{if} \(I_{\pi_{sym}}(s_t,a_t)=1\) \textbf{else} \(1-\eta\)}\;
\textcolor{red}{\(\pi_{ref}(a_t\mid s_t) \leftarrow \frac{w_{sym}(a_t\mid s_t)}{\sum_{a'} w_{sym}(a'\mid s_t)}\)}\;

\textcolor{red}{\(\rho_t^{sym} \leftarrow \frac{\pi_\theta(a_t\mid s_t)}{\pi_{ref}(a_t\mid s_t)}\)}\;  
\textcolor{red}{\(\mathcal{L}^{sym} \leftarrow \mathbb{E}_t\!\left[ \min\left( \rho_t^{sym} \cdot \hat{A}_t,\, \text{clip}(\rho_t^{sym}, 1-\epsilon_{\mathrm{clip}}, 1+\epsilon_{\mathrm{clip}}) \cdot \hat{A}_t \right) \right]\)\;}

\(L \leftarrow -\mathcal{L}^{CLIP} + c_1 \mathcal{L}^{VF} - c_2 \mathcal{H}(\pi_\theta(\cdot\mid s_t)) \textcolor{red}{- \Theta_t \cdot \mathcal{L}^{sym}}\)\;

\(\theta \leftarrow \theta - \alpha \nabla_\theta L\)\;
\end{algorithm2e}


\section{Experiments}
\label{sec:experiments}
\begin{figure*}
    \centering
    \subfigure[DoorKey $8\times8$ - 1 key\label{fig:doorkey}]{%
        \includegraphics[width=0.32\textwidth]{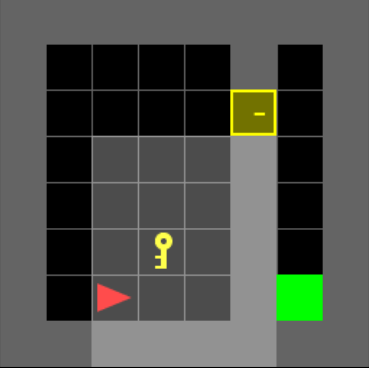}%
    }\hfill
    \subfigure[OfficeWorld\label{fig:office}]{%
        \includegraphics[width=0.32\textwidth]{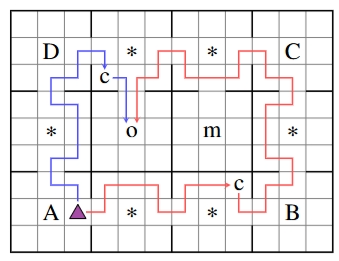}%
    }\hfill
    \subfigure[WaterWorld\label{fig:water}]{%
        \includegraphics[width=0.32\textwidth]{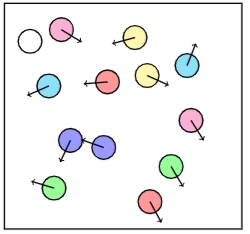}%
    }
\caption{Benchmark domains \label{fig:domains}}
\end{figure*}
We evaluate our methodologies on three benchmarks\footnote{Domain formalizations and symbolic policies are in Appendix \ref{apd:domains}.} (Figure~\ref{fig:domains}): \emph{DoorKey} from \cite{MinigridMiniworld23} and \emph{OfficeWorld} and \emph{WaterWorld} from \cite{icarte2022reward}. \emph{DoorKey} is a partially observable grid navigation task where the agent must retrieve a key that matches the door color before reaching the final goal. \emph{OfficeWorld} takes place in a discrete grid containing labeled locations such as coffee, mail, plants, an office, and various rooms; the agent must navigate through a sequence of sub-goals, while avoiding the plants. \emph{WaterWorld} is a continuous 2D box containing differently colored balls that move at constant velocities and bounce off walls; the agent controls a white ball and must change its velocity to touch specific color sequences in order, while avoiding out-of-sequence collisions. All settings feature sparse rewards and (in some variants) long planning horizons, making them suitable to test whether symbolic action preferences can improve exploration and policy learning.

We compare against vanilla PPO and an RM baseline \citep{icarte2022reward}. In our RM setup, rewards are shaped using (i) a small action-level bonus/penalty depending on whether the chosen action is entailed by \(\pi_{sym}\), and (ii) additional rewards/penalties for progressing or regressing along the states of the RM automaton equivalent to \(\pi_{sym}\).
We tune the main PPO hyperparameters only on the smallest domain instances, and keep them fixed when moving to larger and more complex instances, except for the episode length and the number of parallel environments to accommodate longer horizons. The hyperparameters are reported in Appendix \ref{apd:hyper}.

Crucially, the symbolic policies are not perfect nor complete, but they are induced from the agent's experience in small and simple task instances and then transferred unchanged to harder ones (see Appendix \ref{apd:domains}).
In particular, in \emph{DoorKey} we induce the symbolic policy on the smallest setting ($8\times8$ with 1 key) and reuse it for all larger variants ($8\times8$ with 2/4 keys and $16\times16$ with 1/2/4 keys). In \emph{OfficeWorld}, rules induced in \emph{DeliverCoffee} are reused in \emph{DeliverCoffeeAndMail}, while \emph{PatrolABC} uses its own policy including guidance for rooms $A$, $B$, and $C$. In \emph{WaterWorld}, we induce the symbolic policy on \emph{RedGreen} and reuse the same policy in \emph{RedGreenAndBlueCyan} and \emph{RedGreenAndBlueCyanAndMagentaYellow}.
In the charts, learning curves include \textsc{H-PPO-Product} with \(\varepsilon_r = 0.4\) and \( \varepsilon_f = 0 \) (\textcolor[HTML]{0173B2}{blue}), \textsc{H-PPO-SymLoss} with linearly decaying \(\Theta = \Theta_t = \max(\Theta_i - t \cdot \Theta_r, \Theta_f)\), where \(\Theta_r = 0.4\) and \( \Theta_f = 0 \) (\textcolor[HTML]{CC78BC}{pink}), PPO (\textcolor[HTML]{ED9C0E}{yellow}) and PPO RM (\textcolor[HTML]{029E73}{green}). For the RM baseline, we exclude the additional shaping reward from the plotted return for a fair comparison. We further report an ablation study on the symbolic weight \(\Theta\) for \textsc{H-PPO-SymLoss} in Appendix~\ref{apd:theta} and on \(\varepsilon_f\) for \textsc{H-PPO-Product} in Appendix~\ref{apd:epsilon}.

\subsection{DoorKey Results}
\begin{figure*}
    \centering
    \subfigure[{\footnotesize DoorKey $8\times 8$ - 1 key\label{fig:8x8_1}}]{%
        \includegraphics[width=0.32\linewidth]{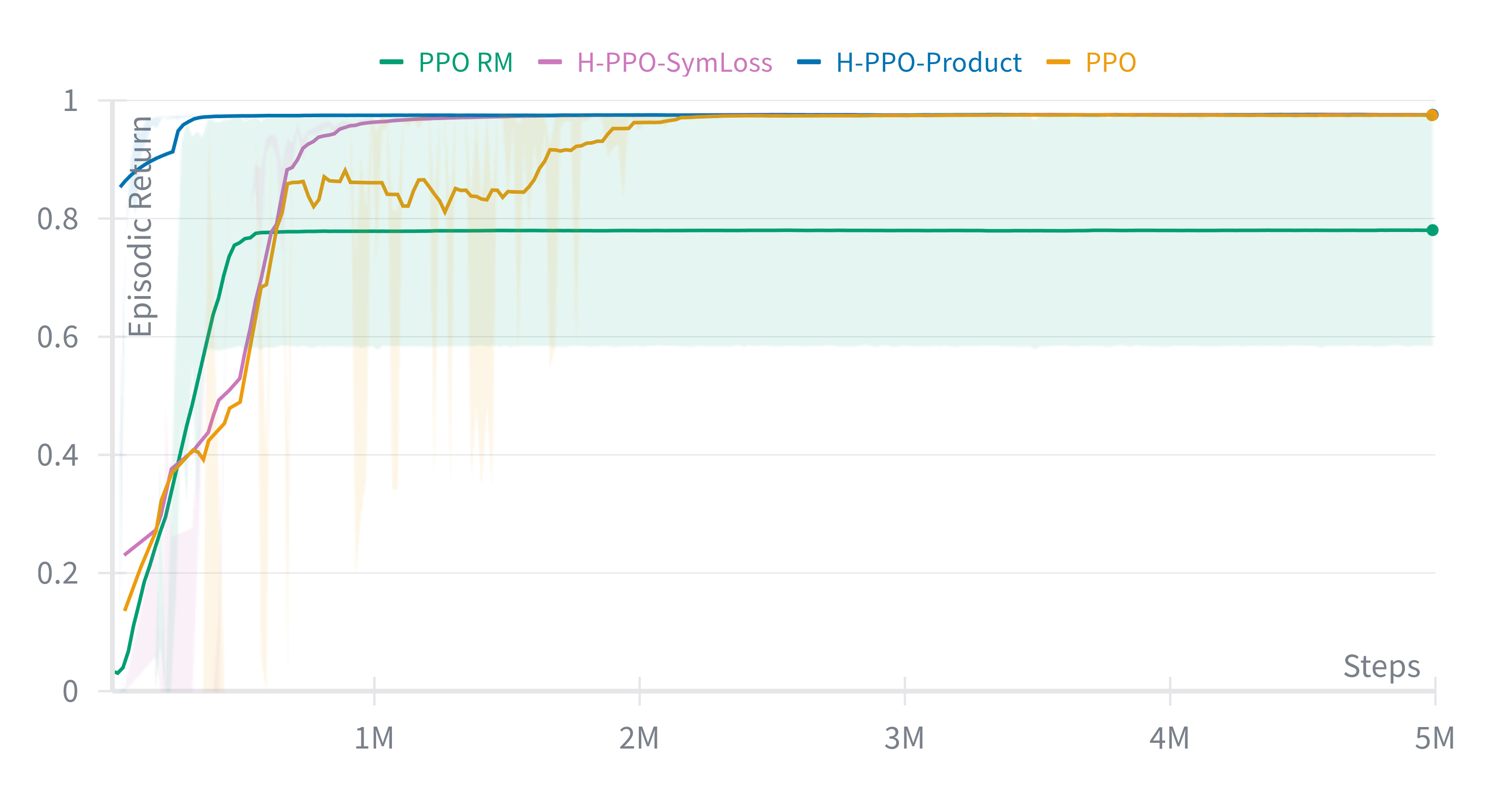}%
    }\hfill
    \subfigure[{\footnotesize DoorKey $8\times 8$ - 2 keys\label{fig:8x8_2}}]{%
        \includegraphics[width=0.32\linewidth]{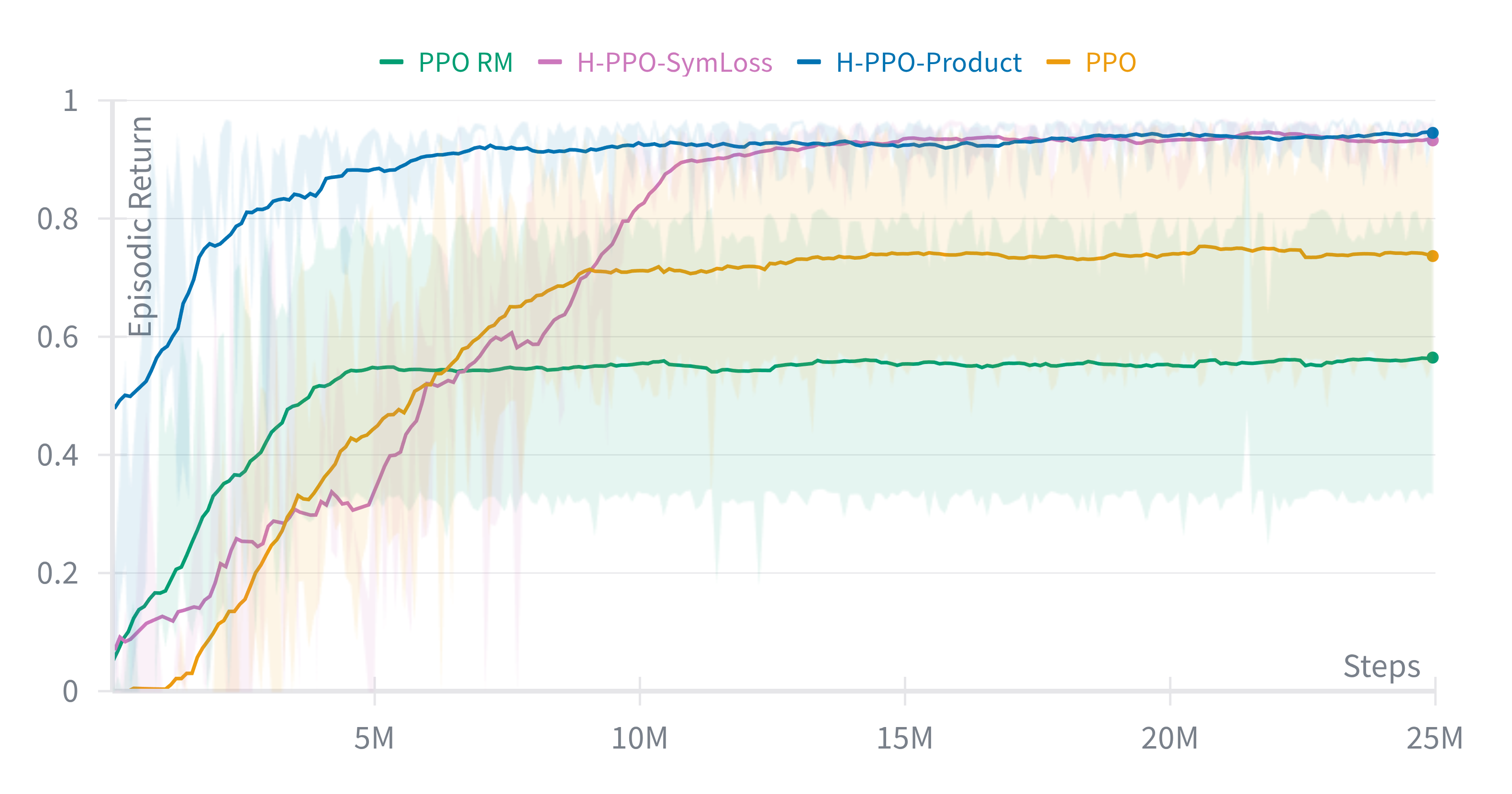}%
    }\hfill
    \subfigure[{\footnotesize DoorKey $8\times 8$ - 4 keys\label{fig:8x8_4}}]{%
        \includegraphics[width=0.32\linewidth]{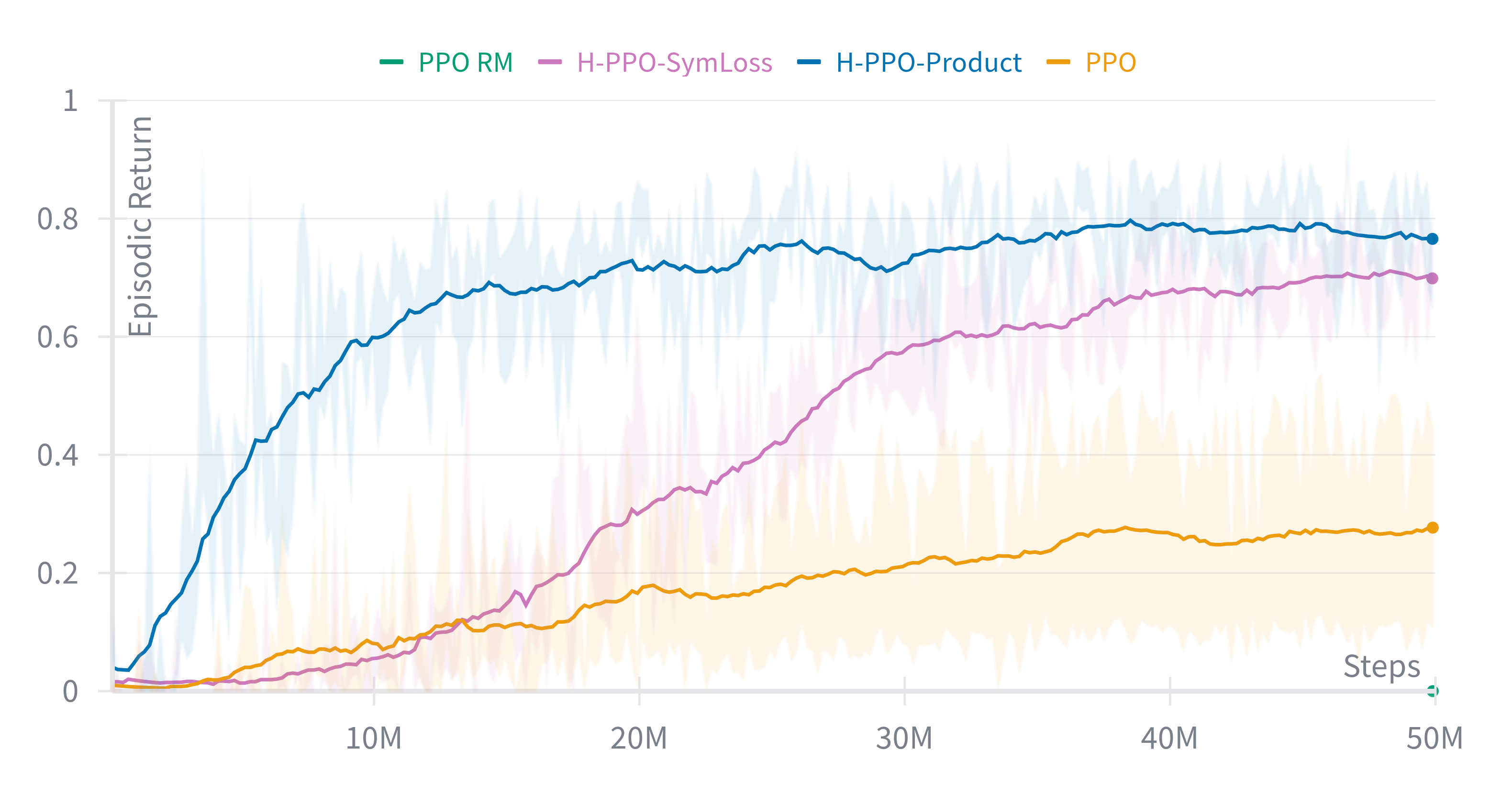}%
    }\\[1em] 
    \subfigure[{\footnotesize DoorKey $16\times 16$ - 1 key\label{fig:16x16_1}}]{%
        \includegraphics[width=0.32\linewidth]{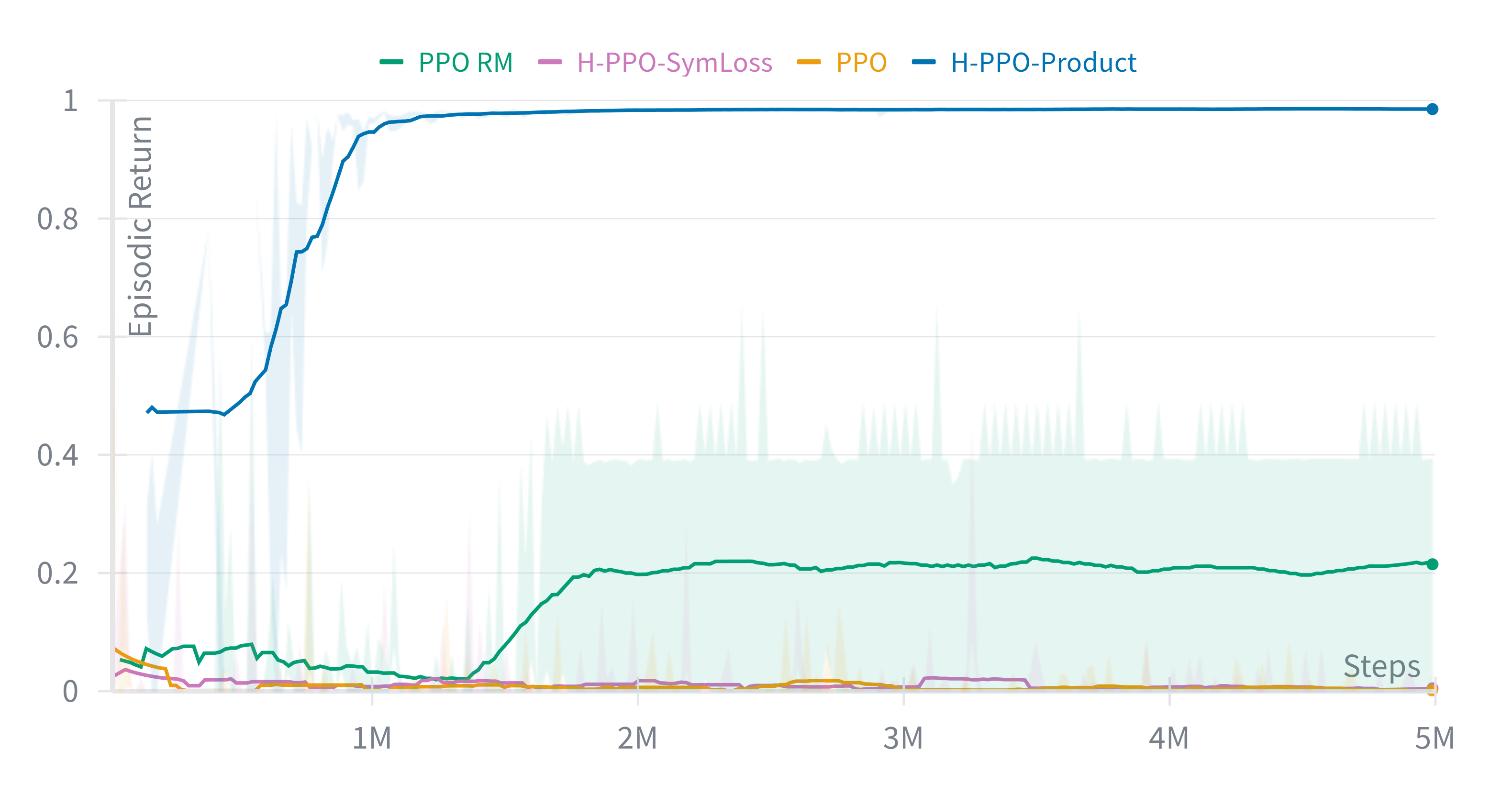}%
    }\hfill
    \subfigure[{\footnotesize DoorKey $16\times 16$ - 2 keys\label{fig:16x16_2}}]{%
        \includegraphics[width=0.32\linewidth]{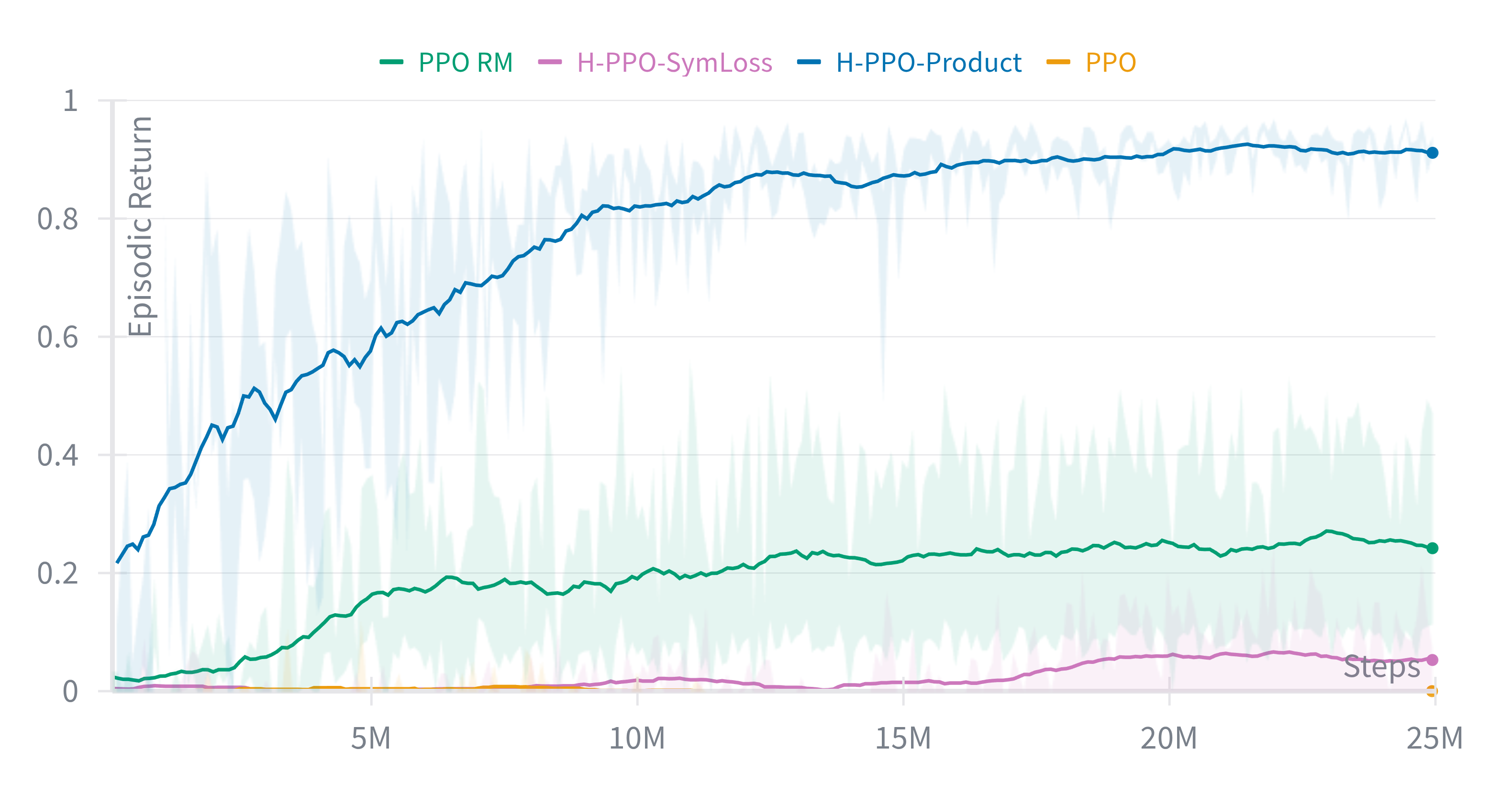}%
    }\hfill
    \subfigure[{\footnotesize DoorKey $16\times 16$ - 4 keys\label{fig:16x16_4}}]{%
        \includegraphics[width=0.32\linewidth]{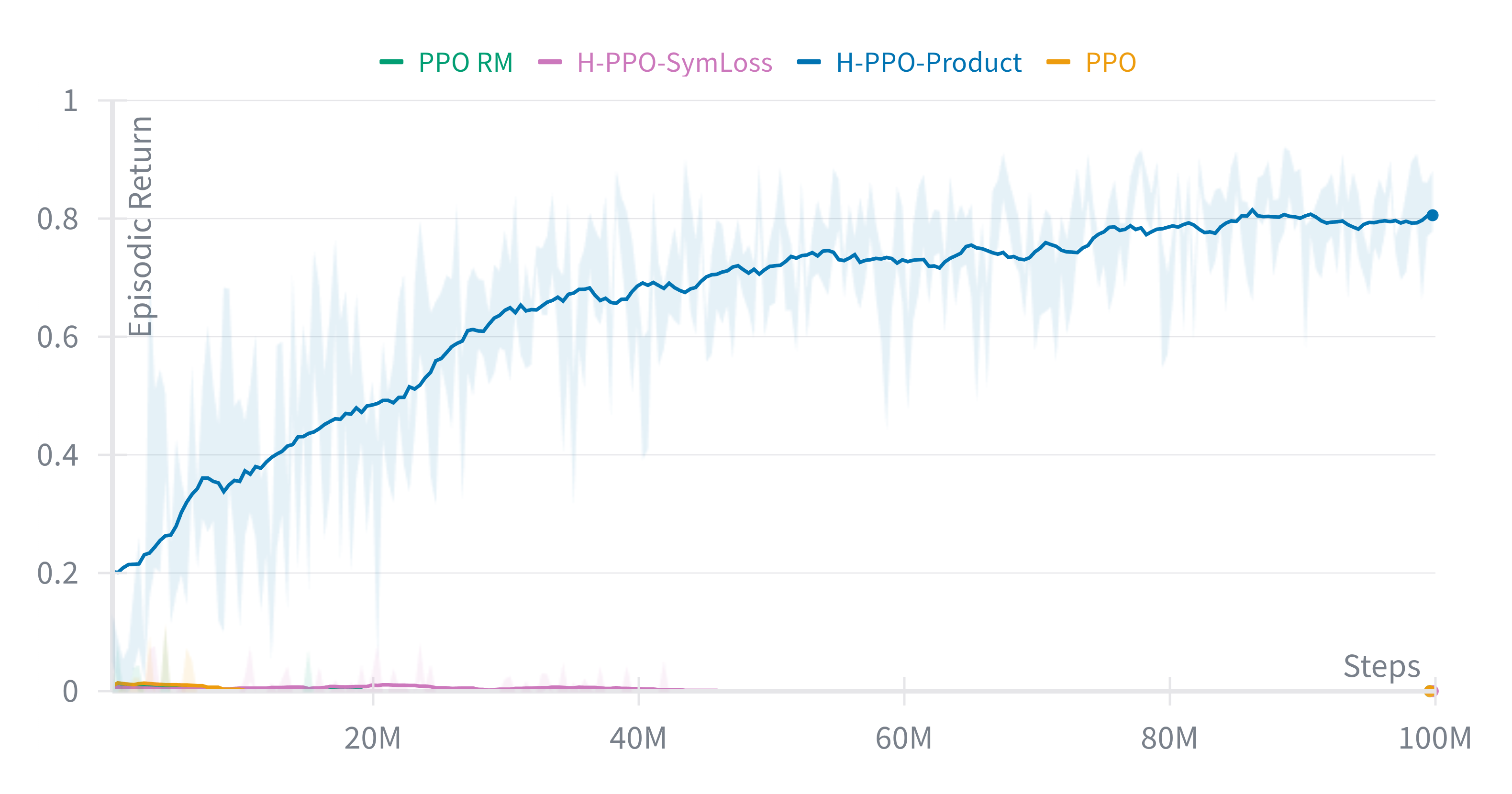}%
    }
    \caption{DoorKey results \label{fig:res_doorkey}}
\end{figure*}
Figure~\ref{fig:res_doorkey} reports the DoorKey results across varying grid sizes
($8\times 8$ and $16\times 16$) and number of keys (1, 2, 4).
PPO hyperparameter tuning was performed in the $8\times 8$ grid with 1 key.

In the $8\times8$ grid (top row), we observe consistent gains from both \textsc{H-PPO-Product} and \textsc{H-PPO-SymLoss} across all key configurations.
With 1 key, all methods converge rapidly, though RMs have more variable performance.
However, as the number of keys increases, our algorithms become increasingly beneficial.
In particular, \textsc{H-PPO-Product} clearly demonstrates its strength in driving exploration, reaching high return with faster growth in the earliest episodes. Meanwhile, \textsc{H-PPO-SymLoss} reaches the same high return, but more slowly.
Interestingly, in the 4-key settings the RM baseline remains at zero return throughout training, so its curve lies on the horizontal axis and is barely visible in the charts, making it the weakest performer.

Scaling to the $16\times16$ grid shows the crucial role of exploration, with a longer planning horizon.
Indeed, \textsc{H-PPO-Product} is the only algorithm reaching near-optimal performance even in the most challenging setting with 4 keys, having the longest planning horizon.

\subsection{OfficeWorld Results}
\begin{figure*}
    \centering
    \subfigure[{\footnotesize DeliverCoffee}]{%
    \includegraphics[width=0.49\linewidth]{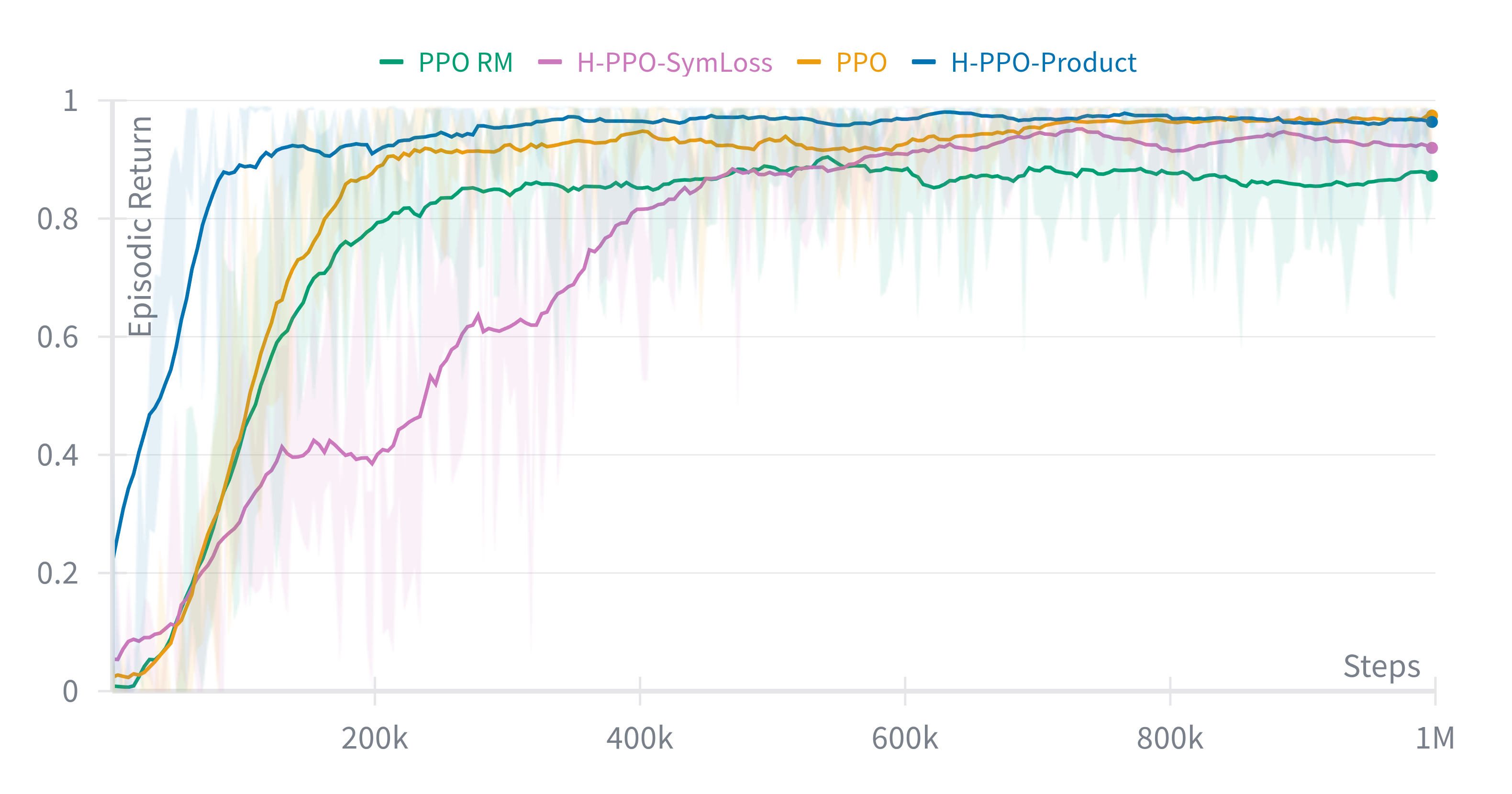}%
    }\hfill
    \subfigure[{\footnotesize PatrolAB}]{%
        \includegraphics[width=0.49\linewidth]{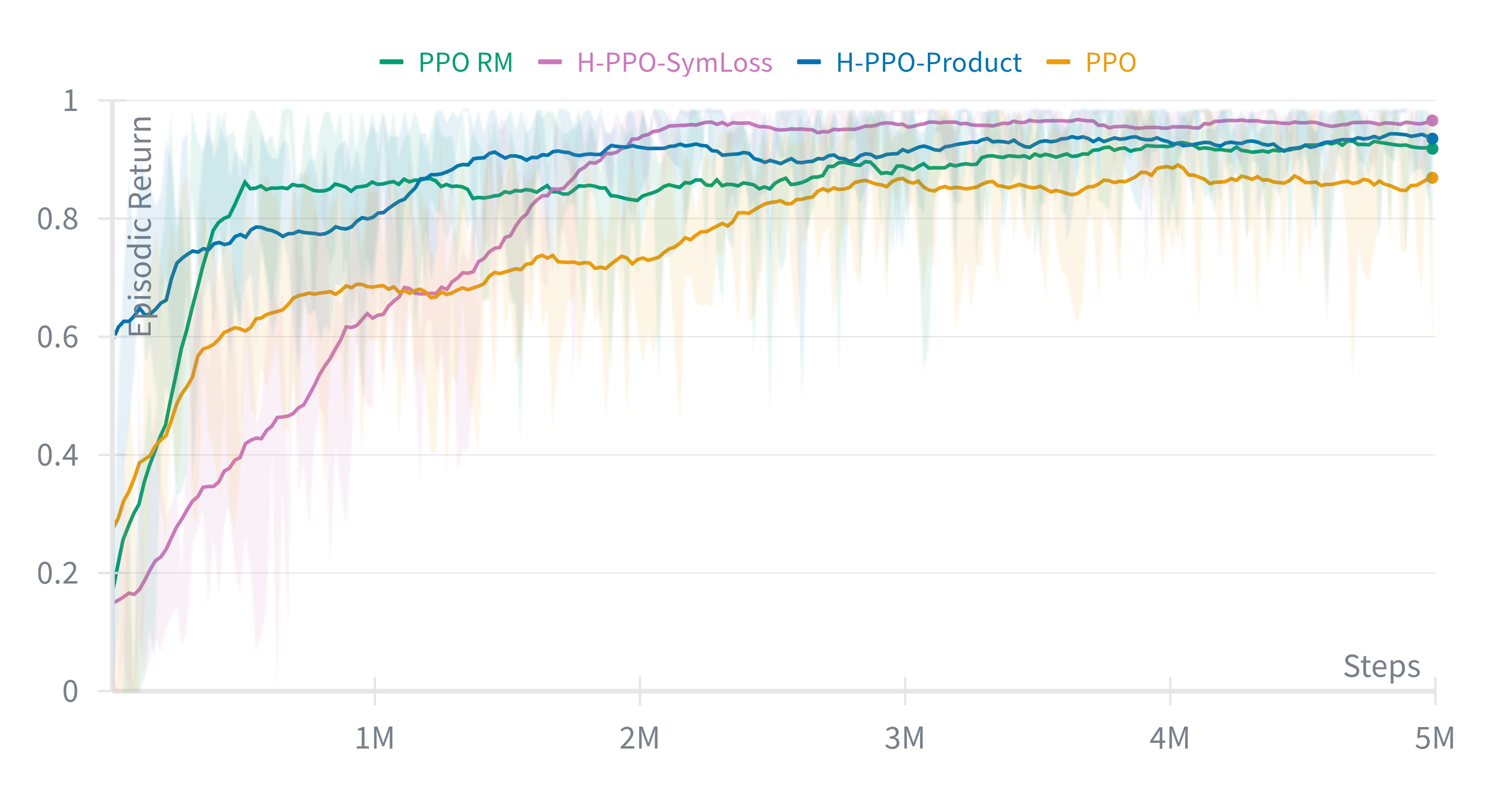}%
    }\\[1em]
    \subfigure[{\footnotesize DeliverCoffeeAndMail}]{%
        \includegraphics[width=0.49\linewidth]{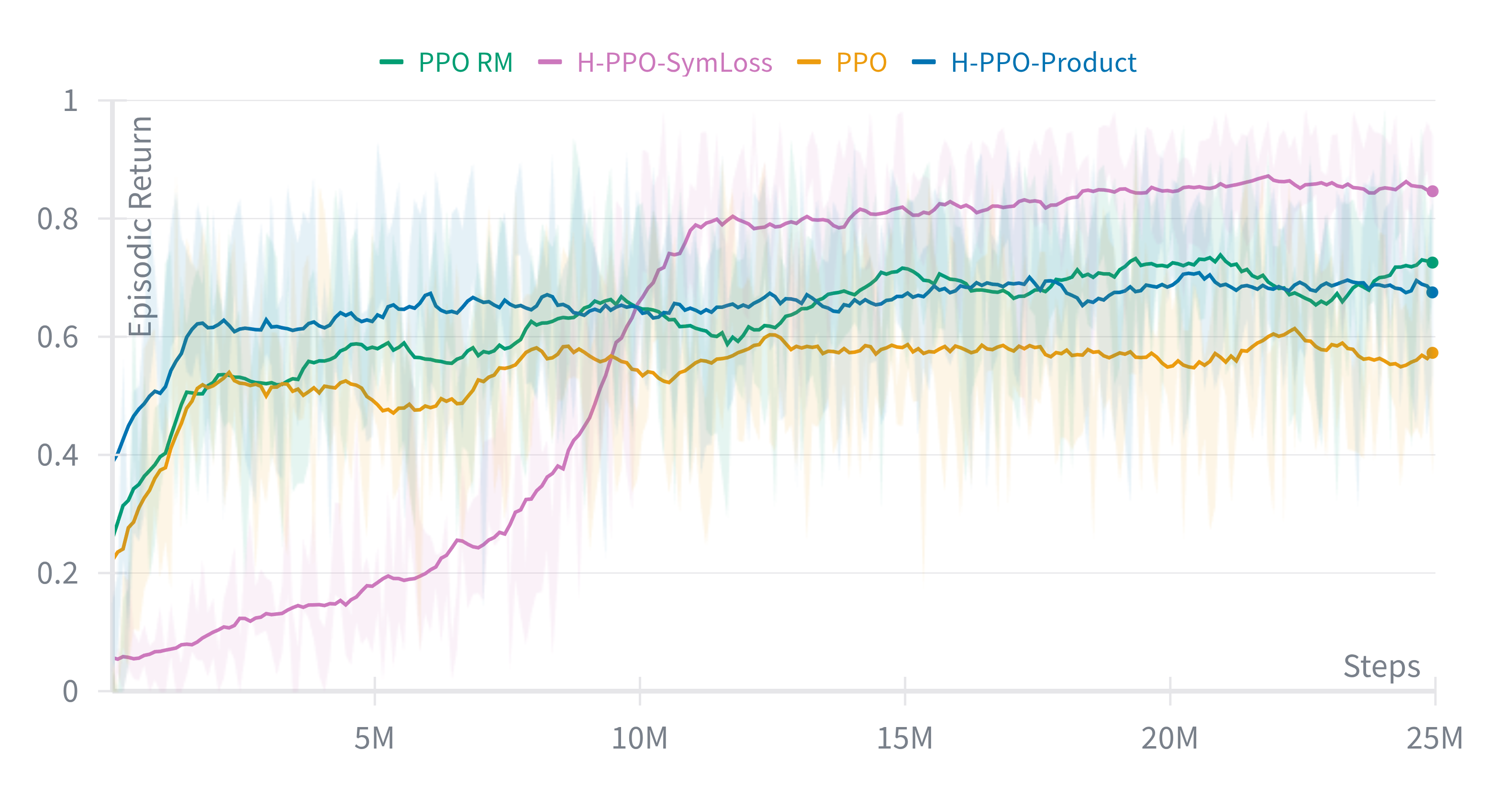}%
    }\hfill
    \subfigure[{\footnotesize PatrolABC}]{%
        \includegraphics[width=0.49\linewidth]{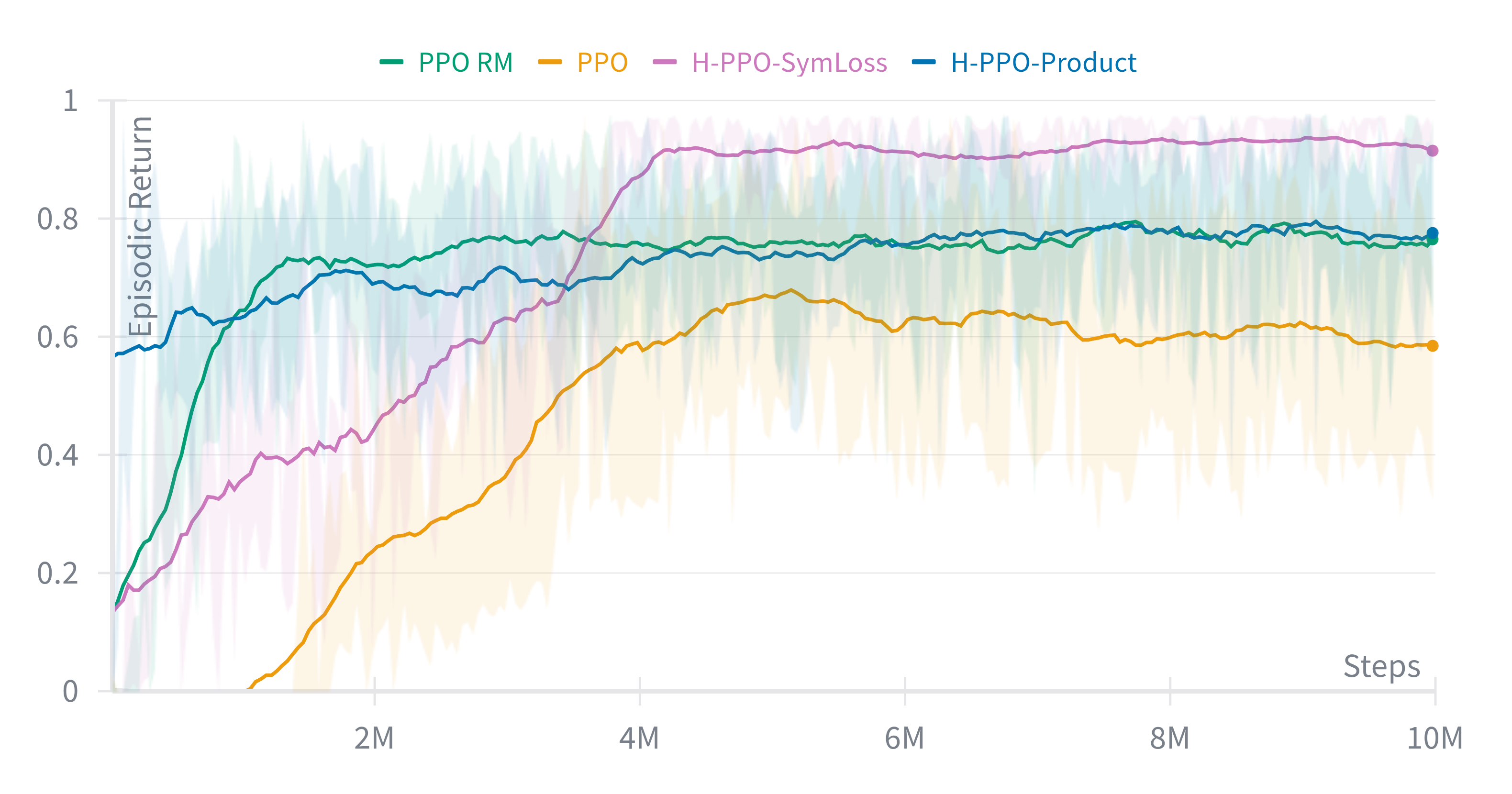}%
    }
    \caption{OfficeWorld results \label{fig:res_officeworld}}
\end{figure*}
Figure~\ref{fig:res_officeworld} reports the OfficeWorld results for \emph{DeliverCoffee}, \emph{PatrolAB}, \emph{DeliverCoffeeAndMail} and \emph{PatrolABC}.

In the simpler tasks \emph{DeliverCoffee} and \emph{PatrolAB}, \textsc{H-PPO-Product} exhibits the fastest rise and reaches near-optimal return early, while PPO and the RM baseline follow closely; \textsc{H-PPO-SymLoss} lags in the first phase but catches up later, ending slightly below the top curves. In \emph{PatrolAB}, the same pattern holds: \textsc{H-PPO-Product} leads the early learning, whereas \textsc{H-PPO-SymLoss} steadily improves and attains the best return at convergence.

Moving to the harder \emph{DeliverCoffeeAndMail} and \emph{PatrolABC} tasks, \textsc{H-PPO-Product} achieves the fastest initial learning due to its exploration bias, but its average return plateaus at a lower level (slightly above $0.65$ in \emph{DeliverCoffeeAndMail}). In contrast, \textsc{H-PPO-SymLoss} improves more slowly at the start (as it provides less exploration pull), but its symbolic regularization strongly shapes the final policy, reaching the highest returns in both tasks. The RM baseline shows moderate performance but remains below \textsc{H-PPO-SymLoss}. These results demonstrate that \textsc{H-PPO-SymLoss} is highly effective at maximizing policy returns, but does not yield a good initial exploration.

\subsection{WaterWorld Results}
\begin{figure*}[h]
    \centering
    \subfigure[{\footnotesize RedGreen\label{fig:redgreen}}]{%
        \includegraphics[width=0.32\linewidth]{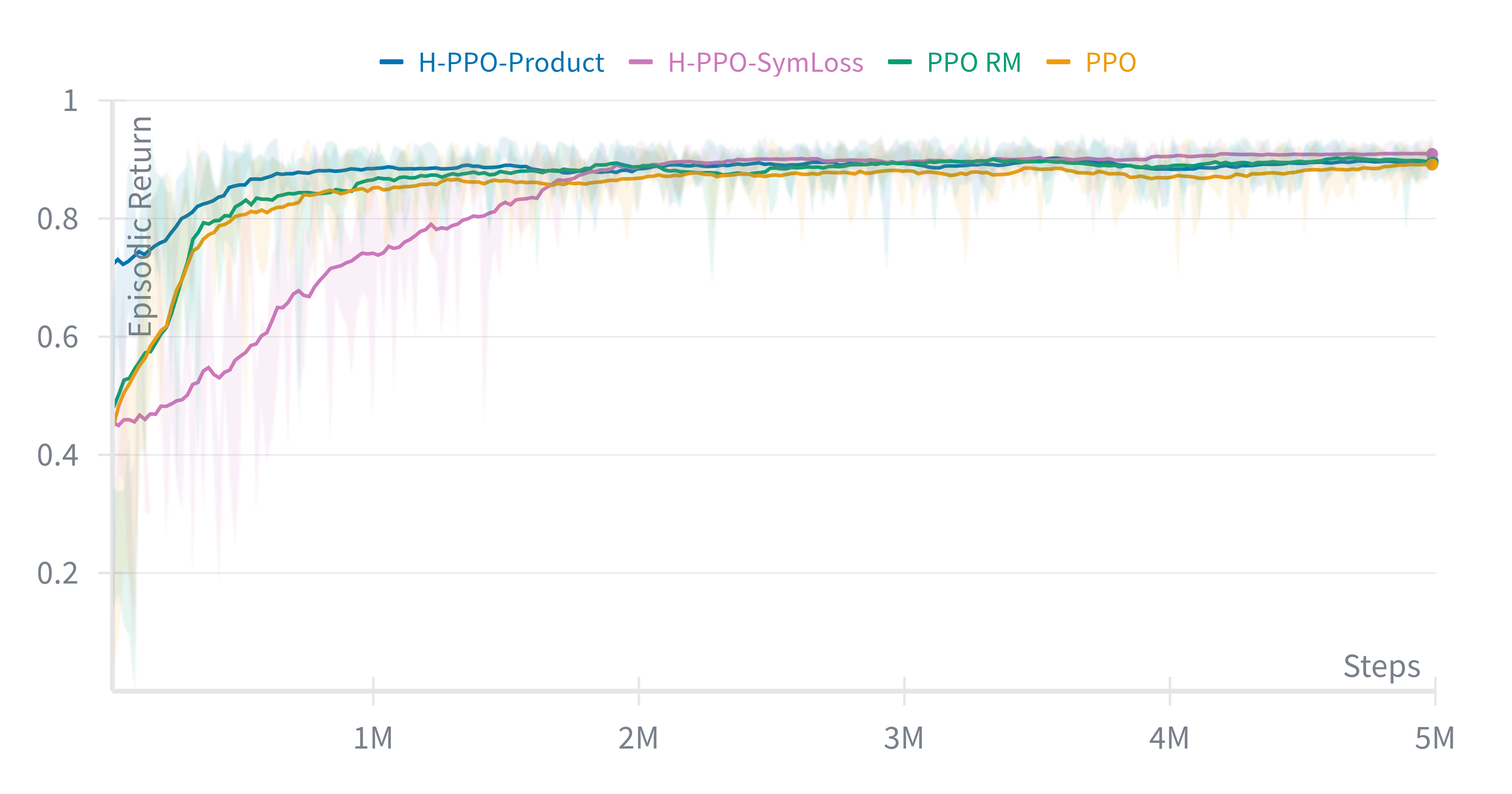}%
    }\hfill
    \subfigure[{\footnotesize RedGreenAndBlueCyan\label{fig:redgreenandbluecyan}}]{%
        \includegraphics[width=0.32\linewidth]{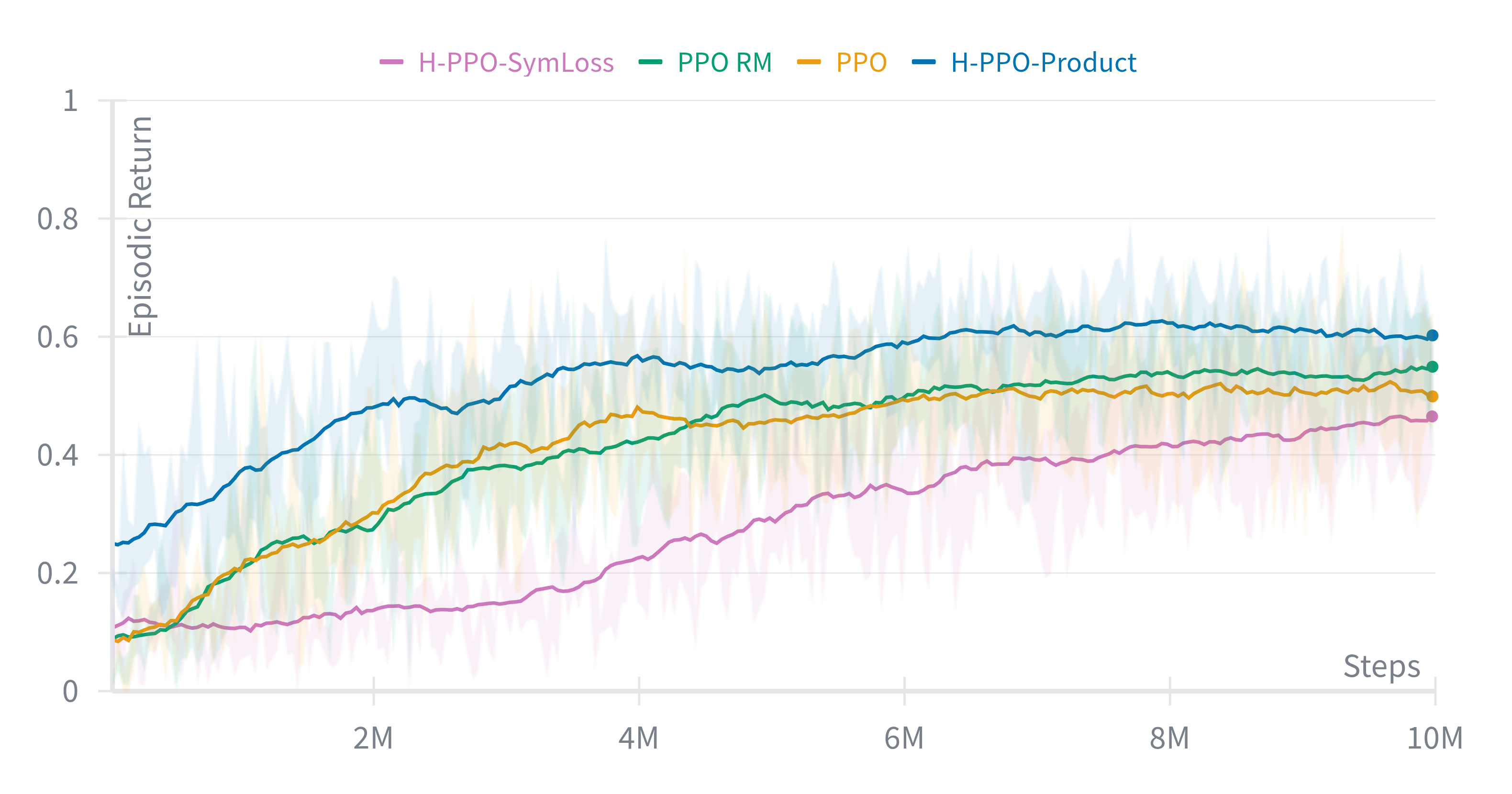}%
    }\hfill
    \subfigure[{\footnotesize RedGreenAndBlueCyan\newline AndMagentaYellow\label{fig:redgreenandbluecyanandmagentayellow}}]{%
        \includegraphics[width=0.32\linewidth]{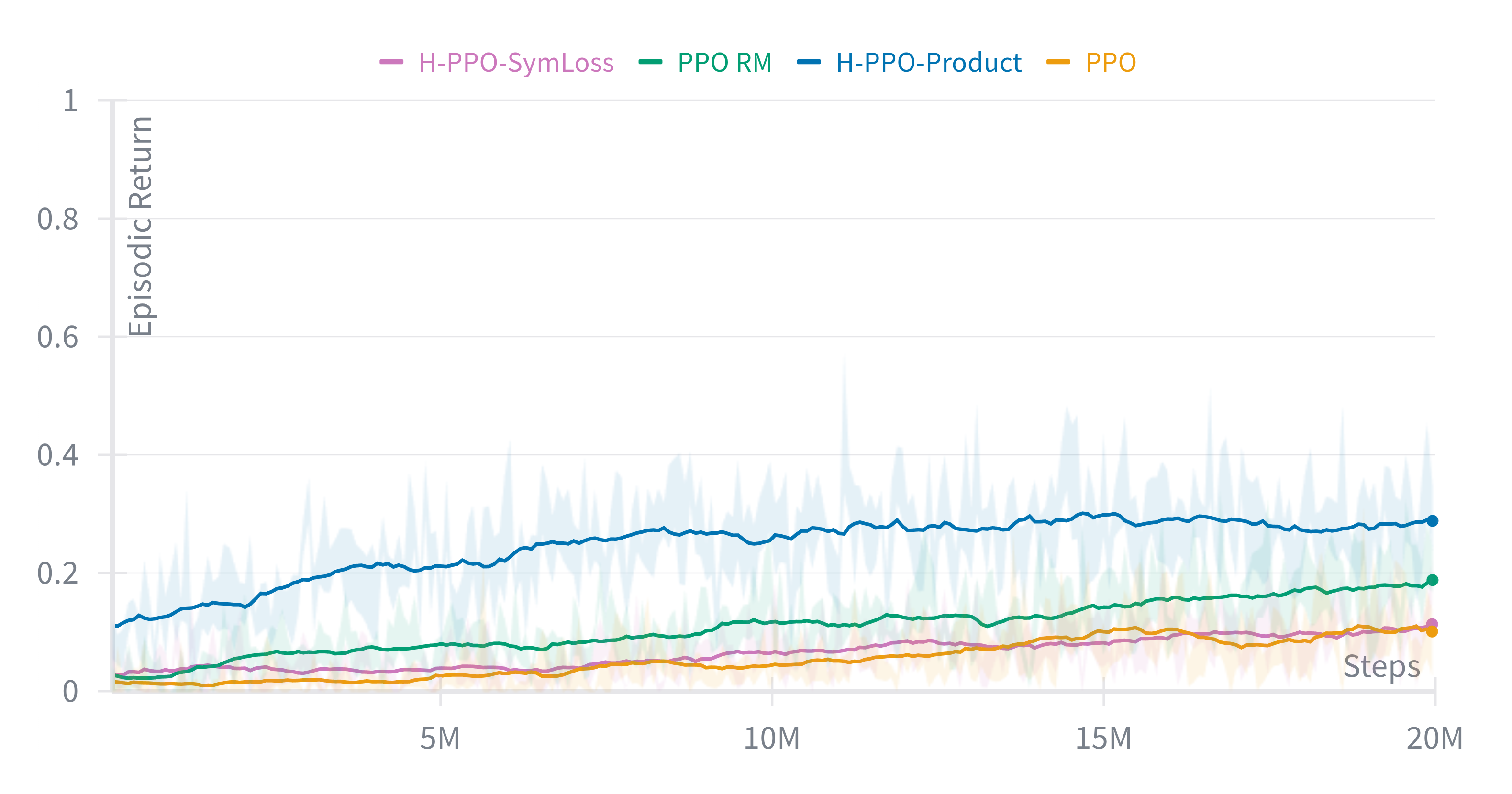}%
    }
    \caption{WaterWorld results \label{fig:res_waterworld}}
\end{figure*}

Figure~\ref{fig:res_waterworld} reports the WaterWorld results for \emph{RedGreen}, \emph{RedGreenAndBlueCyan}, and \emph{RedGreenAndBlueCyanAndMagentaYellow}.

In the \emph{RedGreen} task (where PPO hyperparameters were tuned), all methods easily learn the optimal behavior. 
However, as the tasks become more complex, the differences in performance become clear. In \emph{RedGreenAndBlueCyan}, \textsc{H-PPO-Product} emerges as the most robust approach, actively driving the exploration toward trajectories that follow the required ordering and reaching the highest return close to $0.6$. The RM baseline also performs better than standard PPO. At the same time, \textsc{H-PPO-SymLoss} struggles more than standard PPO, as modifying the policy loss early in training may overly constrain the exploration dynamics.

In the hardest task (\emph{RedGreenAndBlueCyanAndMagentaYellow}), these differences are amplified. As the required sequences become longer, only \textsc{H-PPO-Product} successfully achieves the required exploration to reach higher returns. \textsc{H-PPO-SymLoss} is not effective here, confirming that its primary strength lies in policy consolidation after successes are encountered naturally, rather than in initiating complex exploration sequences out of sparse-reward spaces.

\section{Conclusion}

In this paper, we proposed a neuro-symbolic framework for improving the sample efficiency of PPO in environments characterized by sparse rewards, multiple sub-goals, and long planning horizons.
Our two integration strategies, \textsc{H-PPO-Product} and \textsc{H-PPO-SymLoss}, inject logical rules into the learning process at the level of action selection and policy optimization, respectively, while progressively annealing their influence so that training recovers standard PPO, without hyperparameter re-tuning when scaling to harder instances.

Across three benchmark environments and multiple variants, our methods consistently outperform the baselines, including state-of-the-art reward machines, fulfilling distinct roles. \textsc{H-PPO-Product} proves to be an exploration driver, achieving the fastest convergence and highest final returns in the long, sparse spaces of DoorKey and WaterWorld. Conversely, \textsc{H-PPO-SymLoss} does not significantly aid in early exploration, but acts as a powerful prior for policy regularization; this is clear in OfficeWorld, where it results in the highest returns despite slower initial exploration. Taken together, these results demonstrate that integrating symbolic reasoning directly into the algorithmic structure of DRL offers an effective path towards interpretable, and sample-efficient policy learning and generalization.

Future work will investigate extensions to continuous action spaces, more challenging domains, and the potential for zero-shot generalization in DRL, towards actual real-world application.

\bibliography{nesy2026-sample}
\newpage
\appendix
\section{}\label{appendix}

\subsection{Domains}
\label{apd:domains}
\subsubsection{DoorKey}
DoorKey is a partially observable grid navigation task (Figure \ref{fig:doorkey}) where the agent (red arrow) must retrieve a key, unlock a door of the same color, and then reach the goal cell (green).

The MDP observation covers a $7 \times 7$ area (light gray), with each cell encoded by object type, color and state. The discrete action set has five actions: \texttt{turn left}, \texttt{turn right}, \texttt{move forward}, \texttt{pick up} and \texttt{toggle} (used to open the door).

Rewards are sparse: the agent is rewarded only when it reaches the goal after opening the required door. In that case the reward is $r(s, a) = 1 - 0.9 \cdot \frac{\text{step\_count}}{\text{max\_steps}}$, with $\text{max\_steps} = \text{env\_size} \cdot \text{env\_size} \cdot 10$ as defined by the environment, otherwise the reward is zero.

The symbolic policy contains the logical rules from \cite{hazra2023deep}:
\begin{enumerate}
\item \texttt{pickup(X) $\leftarrow$ key(X), sameColor(X, Y), door(Y), notCarrying} \label{doorkey_1}
\item \texttt{toggle(X) $\leftarrow$ door(X), locked(X), carryingKey(Z), sameColor(X,Z)} \label{doorkey_2}
\item \texttt{goto(X) $\leftarrow$ goal(X), unlocked} \label{doorkey_3}
\end{enumerate}
Rule~\ref{doorkey_1} recommends picking up a key X when it matches the color of a door Y and the agent is not already carrying a key. Rule~\ref{doorkey_2} suggests unlocking a door X when the agent holds a key Z with the same color and the door is locked. Rule~\ref{doorkey_3} then directs the agent toward the goal X once the door is unlocked. Unlike the first two, this rule is not tied to a single low-level action and is treated as a high-level navigation directive.

\subsubsection{OfficeWorld}
OfficeWorld (Figure \ref{fig:office}) is a grid environment with a rectangular layout and static labeled locations such as mail ($m$), coffee ($c$), office ($o$) and decorative plants ($*$) that serve as obstacles. Extra rooms labeled $A$ through $D$ appear depending on the task variant. The agent (red arrow) moves deterministically with four actions: \texttt{up}, \texttt{down}, \texttt{left}, \texttt{right}.

The observation space includes the agent's grid position, whether it is holding coffee, whether it is holding mail, and the progress through room visitation. 
Rewards are sparse and given only at goal completion: $r(s,a) = 1 - 0.9 \cdot \frac{\text{step\_count}}{\text{max\_steps}}$, with $\text{max\_steps} = \text{env\_width} \cdot \text{env\_height} \cdot 10$ by default, otherwise the reward is zero.

We consider four variants of the general task:
\begin{itemize}
    \item \emph{DeliverCoffeeAndMail}, where the agent must collect both coffee and mail before reaching the office. We use the following rules from \cite{furelos2021induction}:
    \begin{enumerate}
    \item \texttt{goto(X) $\leftarrow$ coffee(X), not HasCoffee, notHittingPlants}
    \item \texttt{goto(X) $\leftarrow$ office(X), HasCoffee, notHittingPlants}
    \end{enumerate}
    These rules cover only the deliver coffee subtask.
    \item \emph{DeliverCoffee}, a simplified variant where the agent only needs to collect the coffee and then reach the office. It uses the same rules as \emph{DeliverCoffeeAndMail}.
    \item \emph{PatrolABC}, where the agent must visit rooms $A$, $B$ and $C$ in order while avoiding plants. The corresponding rules by \cite{furelos2021induction} are:
    \begin{enumerate}
    \item \texttt{goto(X) $\leftarrow$ room\_a(X), not visited\_a, not visited\_b} \\
    \hspace*{5.8em}\texttt{not visited\_c, notHittingPlants}
    \item \texttt{goto(X) $\leftarrow$ room\_b(X), visited\_a, not visited\_b} \\
    \hspace*{5.8em}\texttt{not visited\_c, notHittingPlants}
    \item \texttt{goto(X) $\leftarrow$ room\_c(X), visited\_a, visited\_b,} \\
    \hspace*{5.8em}\texttt{not visited\_c, notHittingPlants}
    \end{enumerate}
    \item \emph{PatrolAB}, which only requires visiting rooms $A$ and $B$ in order while avoiding plants. The symbolic policy follows the same structure as \emph{PatrolABC}, but includes only the rules for rooms $A$ and $B$.
\end{itemize}

\subsubsection{WaterWorld}
WaterWorld (Figure \ref{fig:water}) is a continuous 2D box populated by moving colored balls that bounce off the walls. The agent controls a white ball by changing its velocity with five actions: \texttt{up}, \texttt{down}, \texttt{left}, \texttt{right}, and \texttt{none}. At every step, the observation is the set of colors currently colliding with the agent, and the task is to touch balls in a prescribed color sequence. Rewards are sparse: the agent receives $r(s,a) = 1 - 0.9 \cdot \frac{\text{step\_count}}{\text{max\_steps}}$ only when the sequence(s) are completed, otherwise zero, with $\text{max\_steps} = \text{env\_size} / 2$ by default.

We evaluate three variants where multiple sequences can be interleaved. The logical rules are simple progression constraints over the color-touch events:
\begin{itemize}
    \item \emph{RedGreen}:
    \begin{enumerate}
        \item \texttt{touch(red) $\leftarrow$ not touched\_red, not touched\_green.}
        \item \texttt{touch(green) $\leftarrow$ touched\_red, not touched\_green.}
    \end{enumerate}
    The first rule recommends seeking a red collision until it happens; the second switches to green once red has been touched. They are both not actually tied to a single low-level action and are treated as high-level navigation directives.
    \item \emph{RedGreenAndBlueCyan} and \emph{RedGreenAndBlueCyanAndMagentaYellow}: the symbolic policy is the same as \emph{RedGreen}; the blue$\rightarrow$cyan and magenta$\rightarrow$yellow sequences are unsuggested.
\end{itemize}
\newpage

\subsection{Hyperparameters}
\label{apd:hyper}
\begin{table*}[h]
\centering
\begin{tabular}{lrrrrr}
\hline
Task & total\_timesteps & num\_envs & batch\_size & minibatch\_size \\ \hline
8x8 1 key & \num{5000000} & 4 & 512 & 128 \\
8x8 2 keys & \num{25000000} & 8 & 1024 & 256 \\
8x8 4 keys & \num{50000000} & 16 & 2048 & 512 \\
16x16 1 key & \num{5000000} & 16 & 2048 & 512 \\
16x16 2 keys & \num{25000000} & 32 & 4096 & 1024 \\
16x16 4 keys & \num{100000000} & 64 & 8192 & 2048 \\
\hline
\end{tabular}
\caption{\textbf{Hyperparameters for \emph{DoorKey}}. Fixed across all tasks: learning\_rate=0.0003, num\_steps=128, anneal\_lr=true, gamma=0.99, gae\_lambda=0.95, num\_minibatches=4, update\_epochs=4, norm\_adv=true, clip\_coef=0.2, clip\_vloss=true, ent\_coef=0.01, vf\_coef=0.5, max\_grad\_norm=0.5.}
\label{tab:hyperparams-doorkey}
\end{table*}

\begin{table*}[h]
\centering
\begin{tabular}{lrrrr}
\hline
Task & total\_timesteps & num\_envs & batch\_size & minibatch\_size \\ \hline
DeliverCoffee & \num{1000000} & 8 & 1024 & 256 \\
DeliverCoffeeAndMail & \num{25000000} & 32 & 4096 & 1024 \\
PatrolAB & \num{5000000} & 8 & 1024 & 256 \\
PatrolABC & \num{10000000} & 8 & 1024 & 256 \\
\hline
\end{tabular}
\caption{\textbf{Hyperparameters for \emph{OfficeWorld}}. Fixed across all tasks: learning\_rate=0.0003, num\_steps=128, anneal\_lr=true, gamma=0.99, gae\_lambda=0.95, num\_minibatches=4, update\_epochs=4, norm\_adv=true, clip\_coef=0.2, clip\_vloss=true, ent\_coef=0.01, vf\_coef=0.5, max\_grad\_norm=0.5.}
\label{tab:hyperparams-office}
\end{table*}

\begin{table*}[h]
\centering
\begin{tabular}{lrrrrr}
\hline
Task & total\_timesteps & num\_envs & batch\_size & minibatch\_size \\ \hline
RG & \num{5000000} & 8 & 1024 & 256 \\
RG\&BC & \num{10000000} & 16 & 2048 & 512 \\
RG\&BC\&MY & \num{20000000} & 32 & 4096 & 1024 \\
\hline
\end{tabular}
\caption{\textbf{Hyperparameters for \emph{WaterWorld}}. Task abbreviations: RG = RedGreen, RG\&BC = RedGreenAndBlueCyan, RG\&BC\&MY = RedGreenAndBlueCyanAndMagentaYellow. Fixed across all tasks: learning\_rate=0.0003, num\_steps=128, anneal\_lr=true, gamma=0.99, gae\_lambda=0.95, num\_minibatches=4, update\_epochs=4, norm\_adv=true, clip\_coef=0.2, clip\_vloss=true, ent\_coef=0.01, vf\_coef=0.5, max\_grad\_norm=0.5.}
\label{tab:hyperparams-water}
\end{table*}
\newpage

\subsection{Ablation Study for \texorpdfstring{\(\Theta\)}{Theta} in \textsc{H-PPO-SymLoss}}
\label{apd:theta}
\begin{figure*}[h]
    \centering
    \subfigure[DoorKey $8\times 8$ - 1 key]{%
        \includegraphics[width=0.32\linewidth]{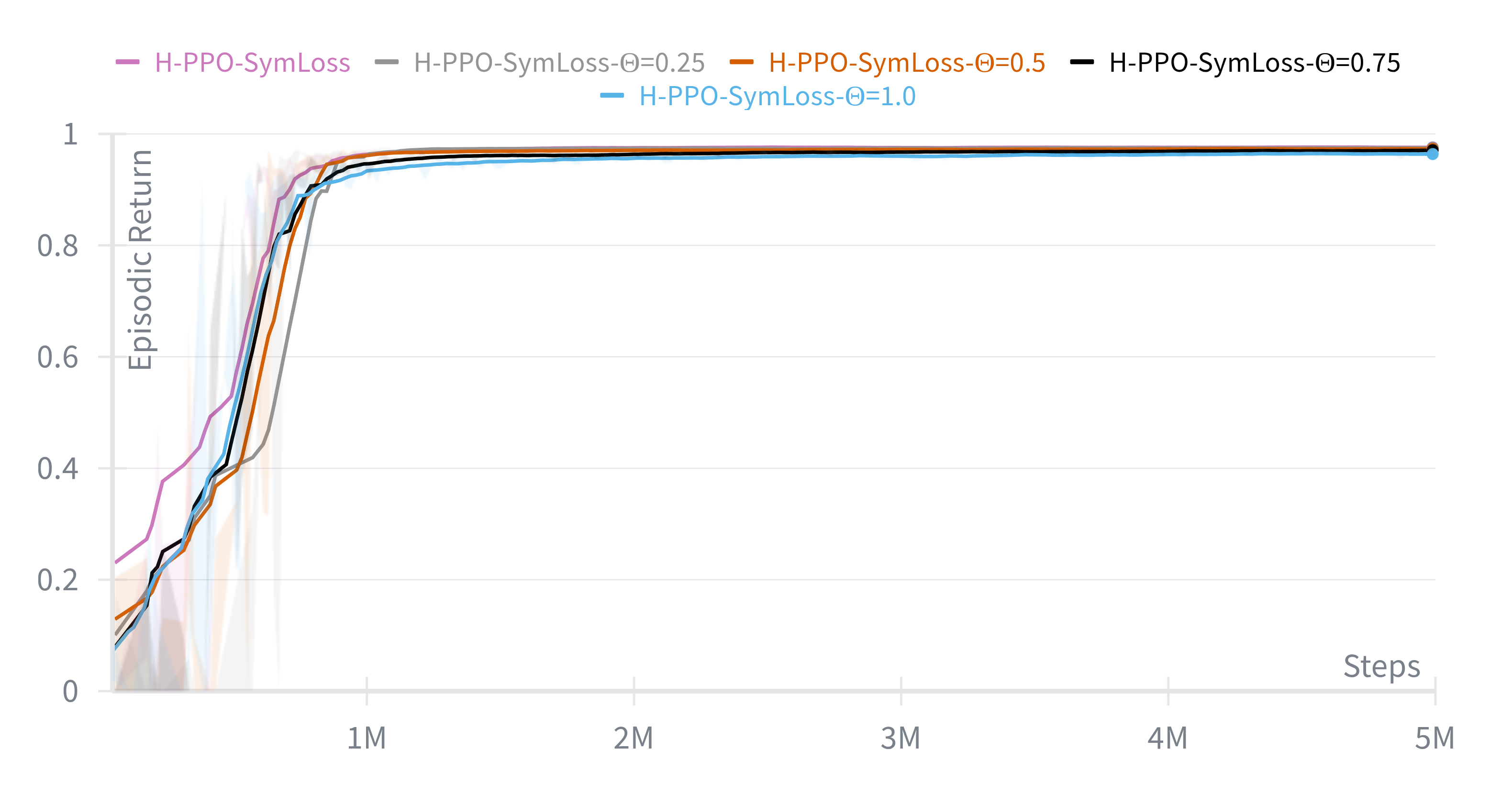}%
    }\hfill
    \subfigure[DoorKey $8\times 8$ - 2 keys]{%
        \includegraphics[width=0.32\linewidth]{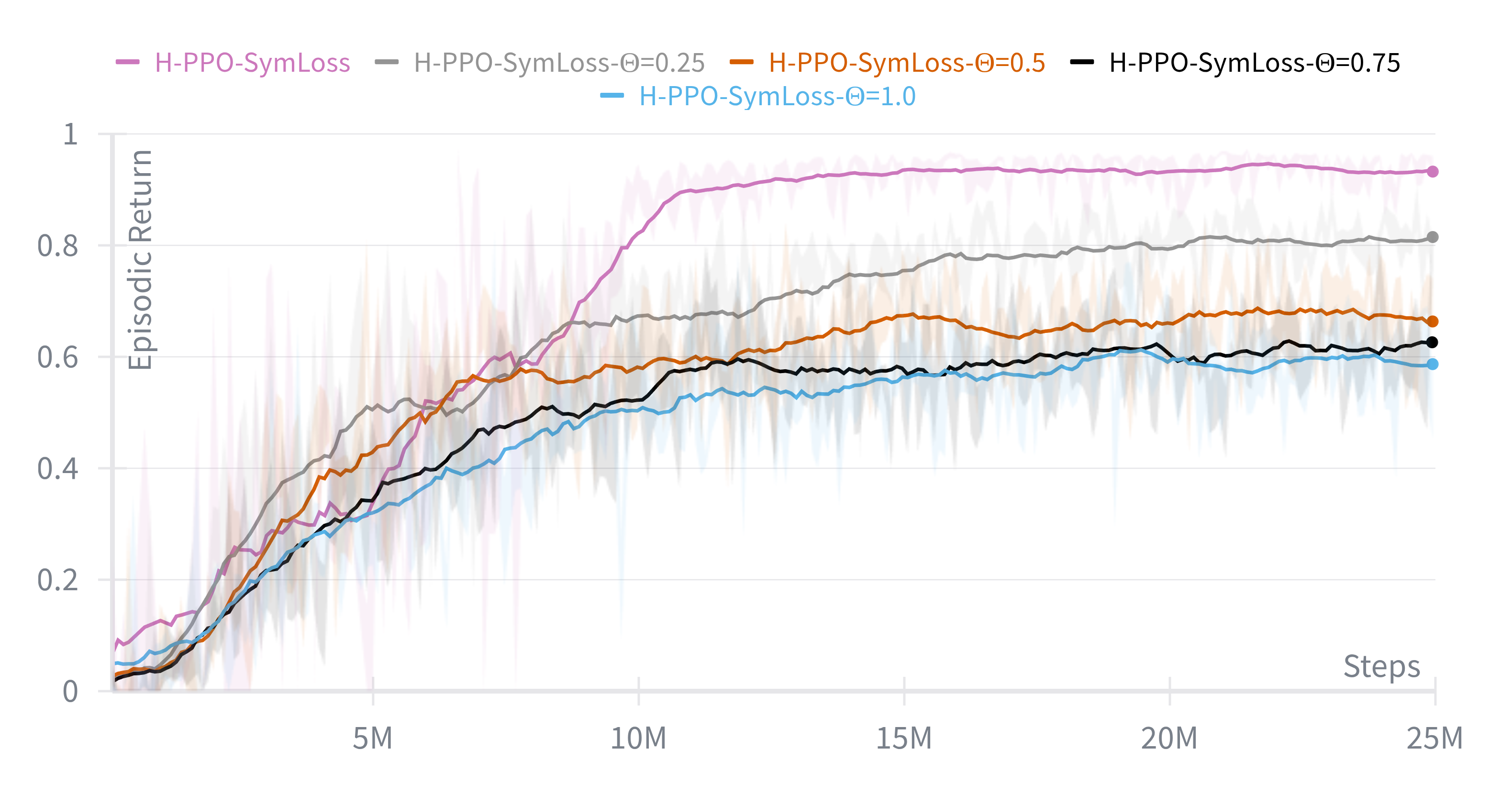}%
    }\hfill
    \subfigure[DoorKey $8\times 8$ - 4 keys]{%
        \includegraphics[width=0.32\linewidth]{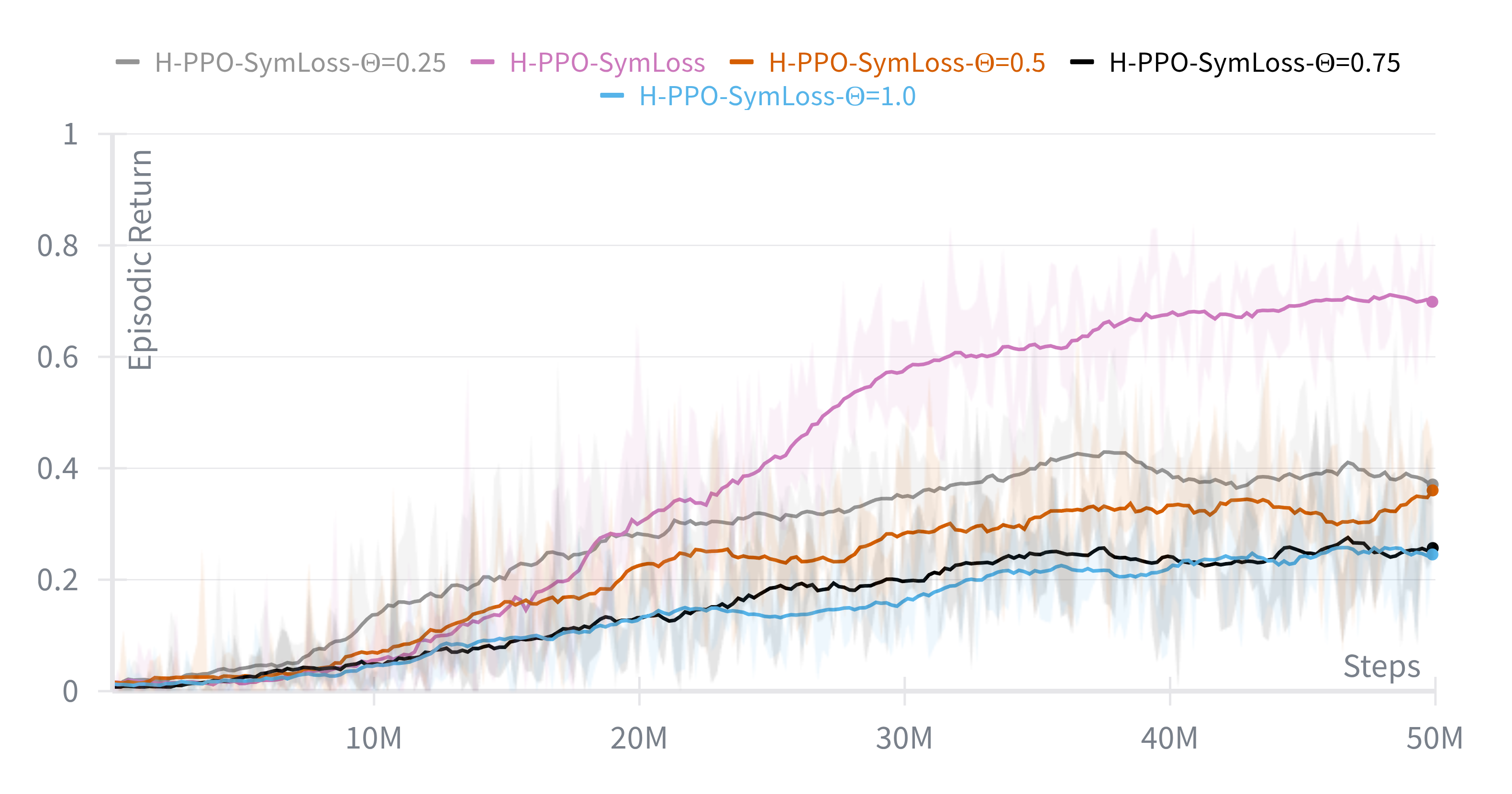}%
    }\\[1em]
    \subfigure[DoorKey $16\times 16$ - 1 key]{%
        \includegraphics[width=0.32\linewidth]{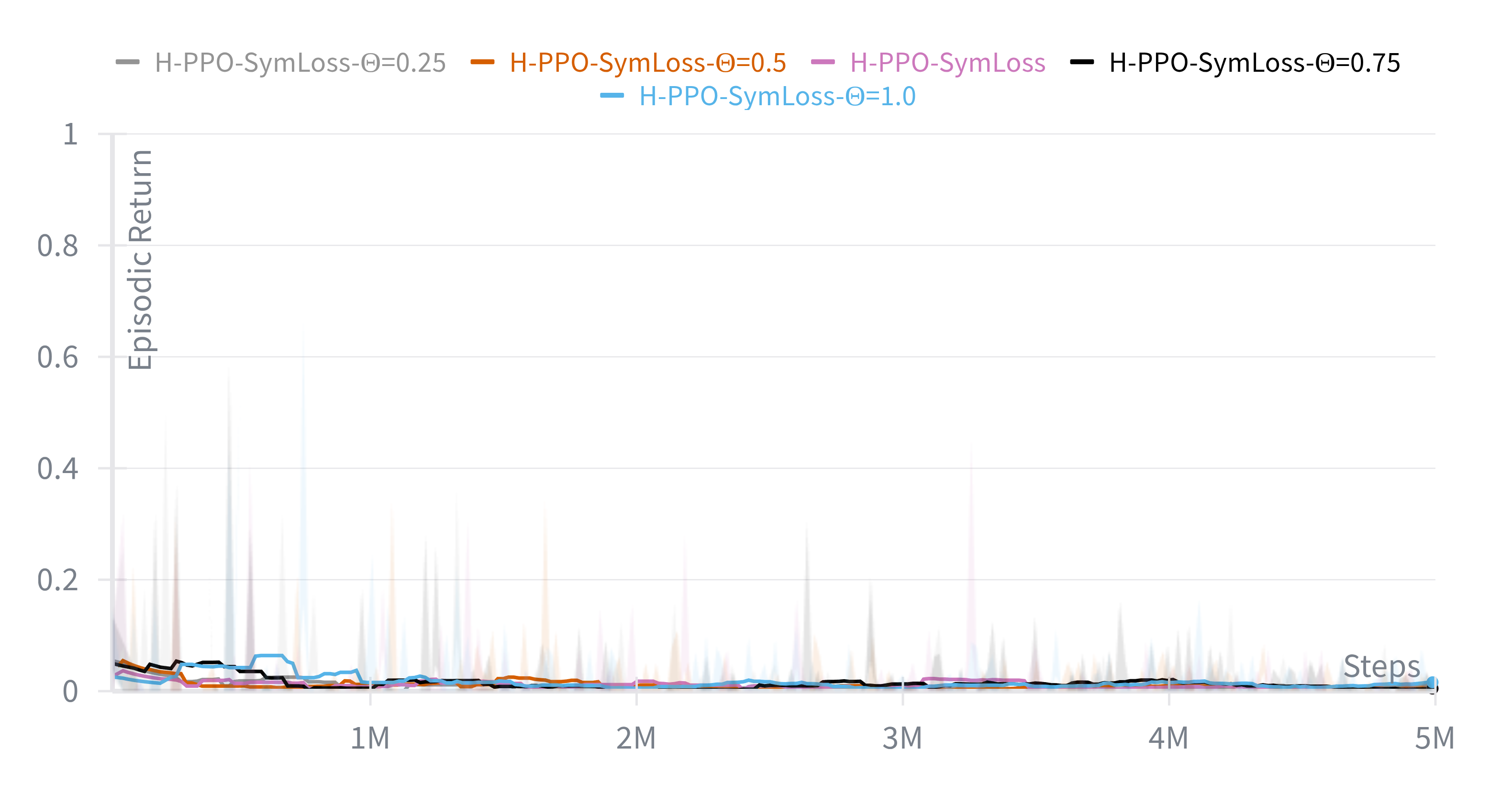}%
    }\hfill
    \subfigure[DoorKey $16\times 16$ - 2 keys]{%
        \includegraphics[width=0.32\linewidth]{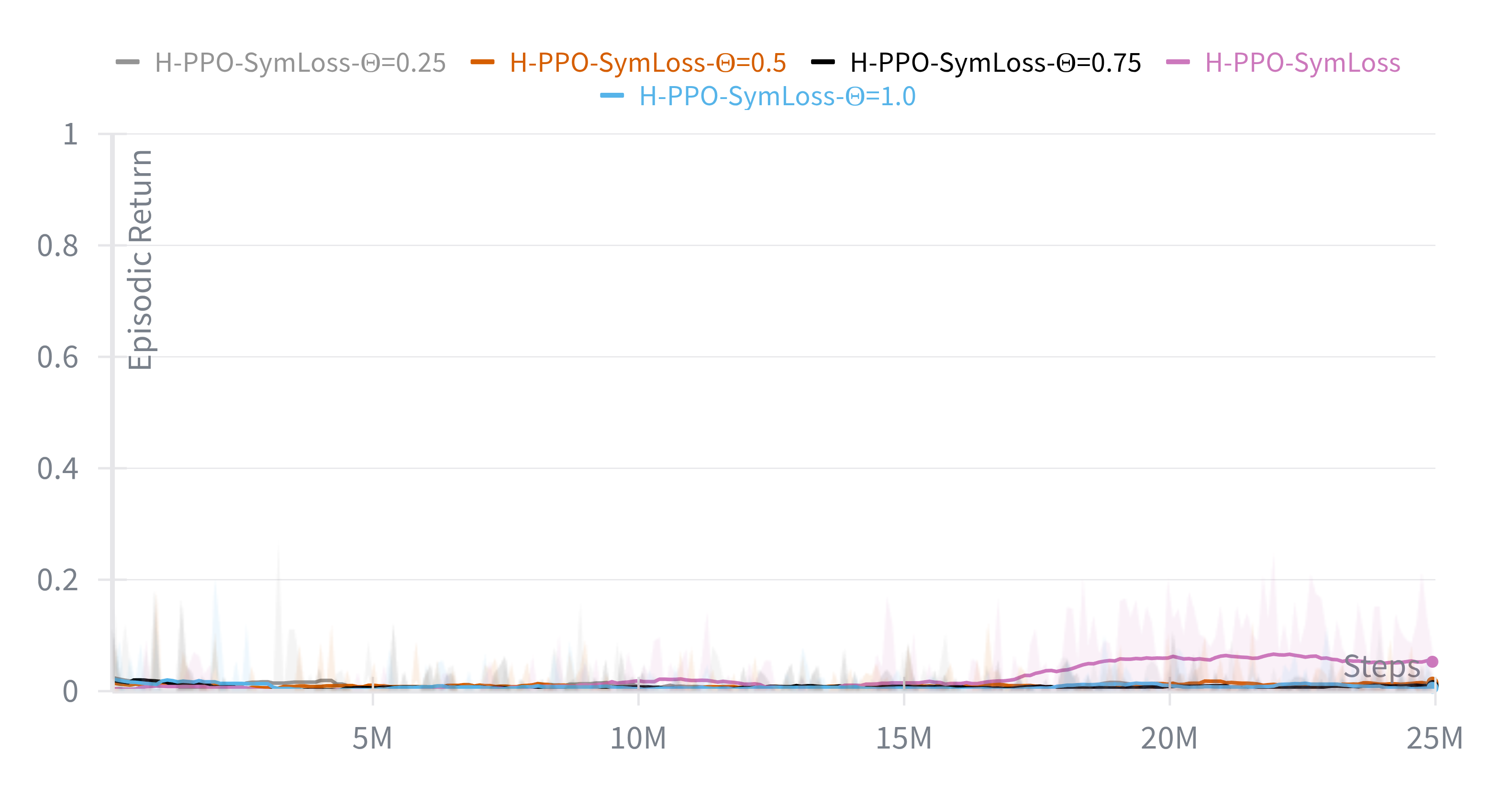}%
    }\hfill
    \subfigure[DoorKey $16\times 16$ - 4 keys]{%
        \includegraphics[width=0.32\linewidth]{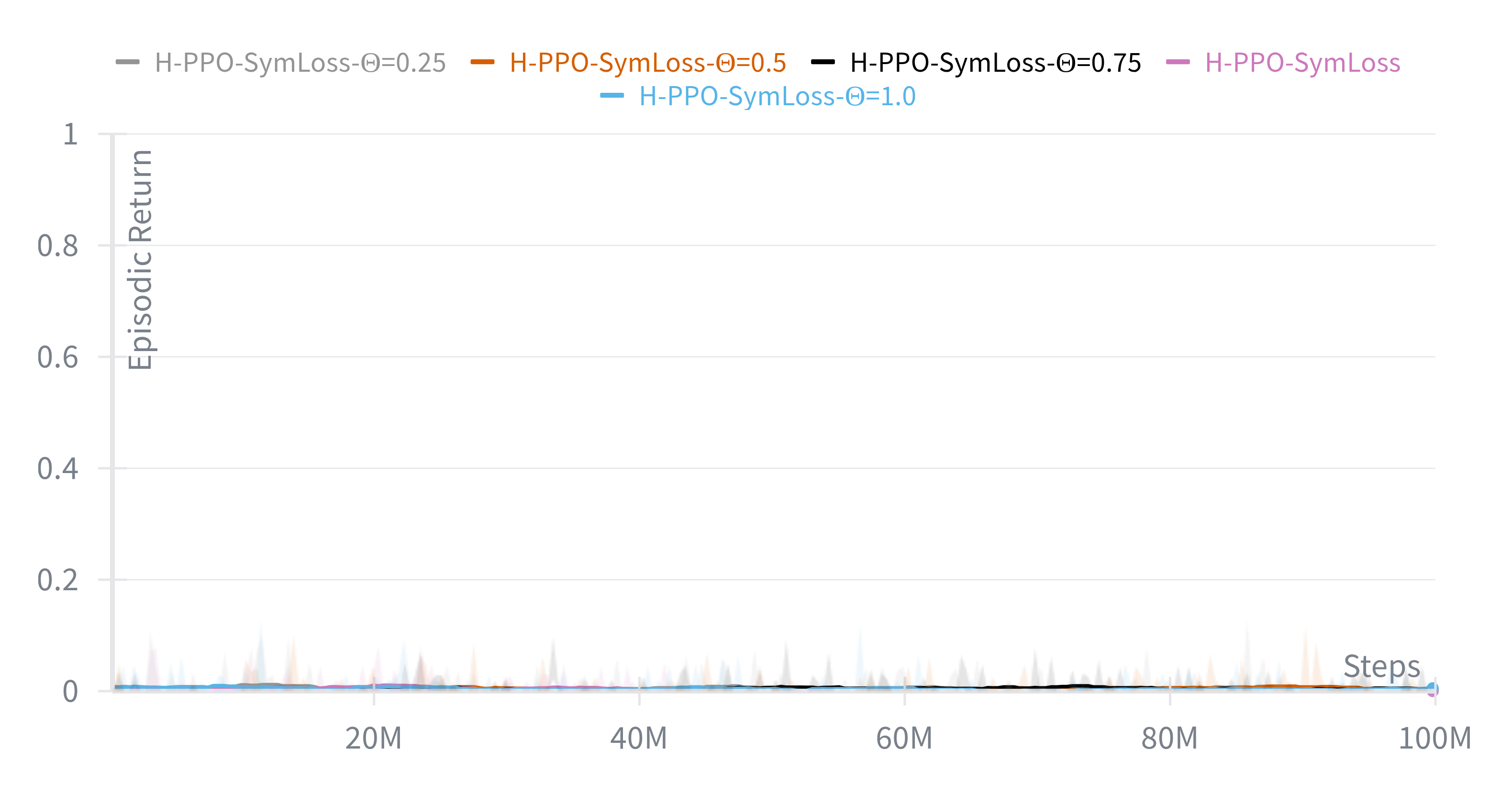}%
    }
    \caption{Ablation charts for \(\Theta\) in DoorKey \label{fig:ablation_theta_doorkey}}
\end{figure*}

\begin{figure*}
    \centering
    \subfigure[DeliverCoffee]{%
        \includegraphics[width=0.49\linewidth]{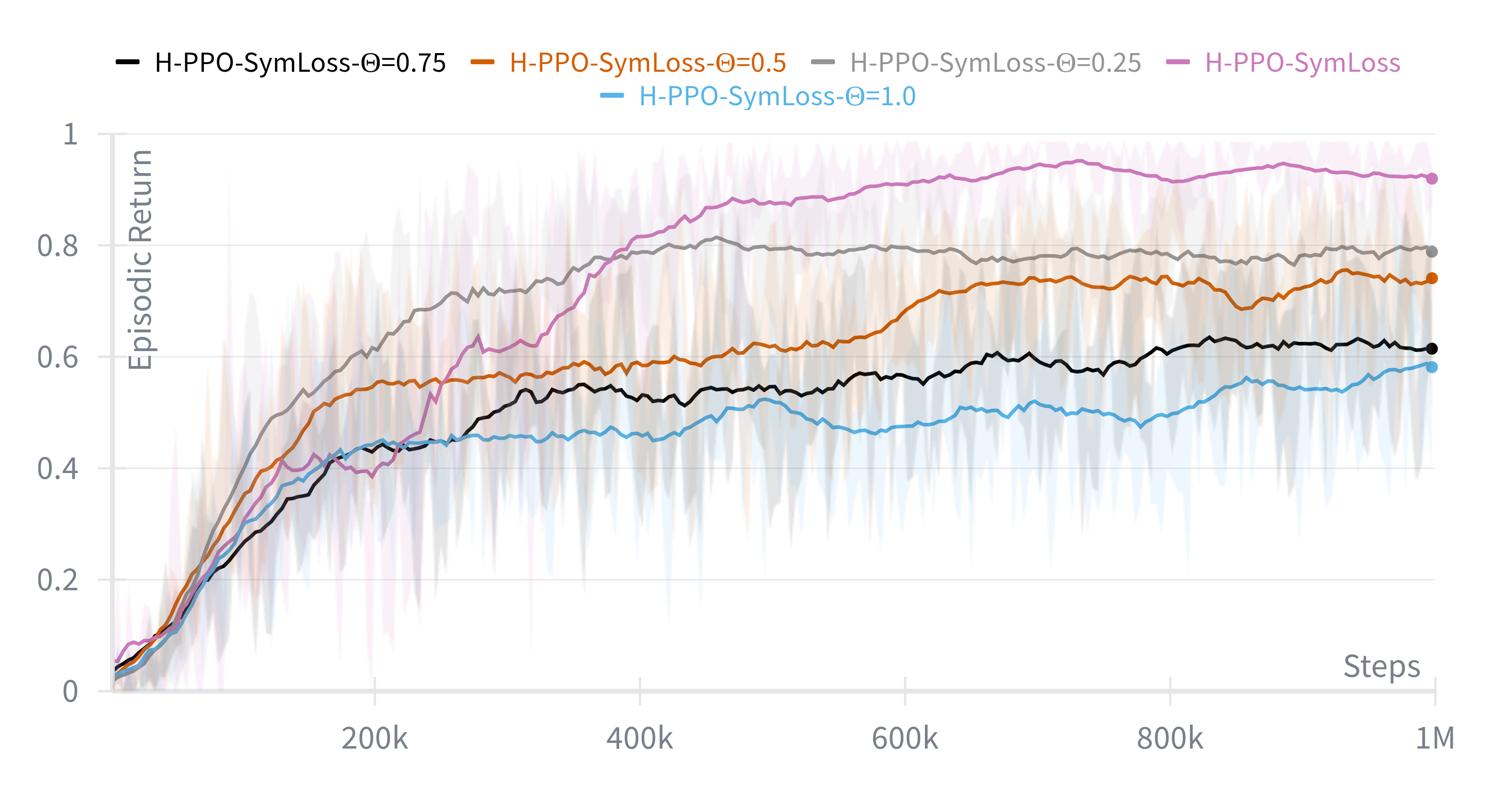}%
    }\hfill
    \subfigure[PatrolAB]{%
        \includegraphics[width=0.49\linewidth]{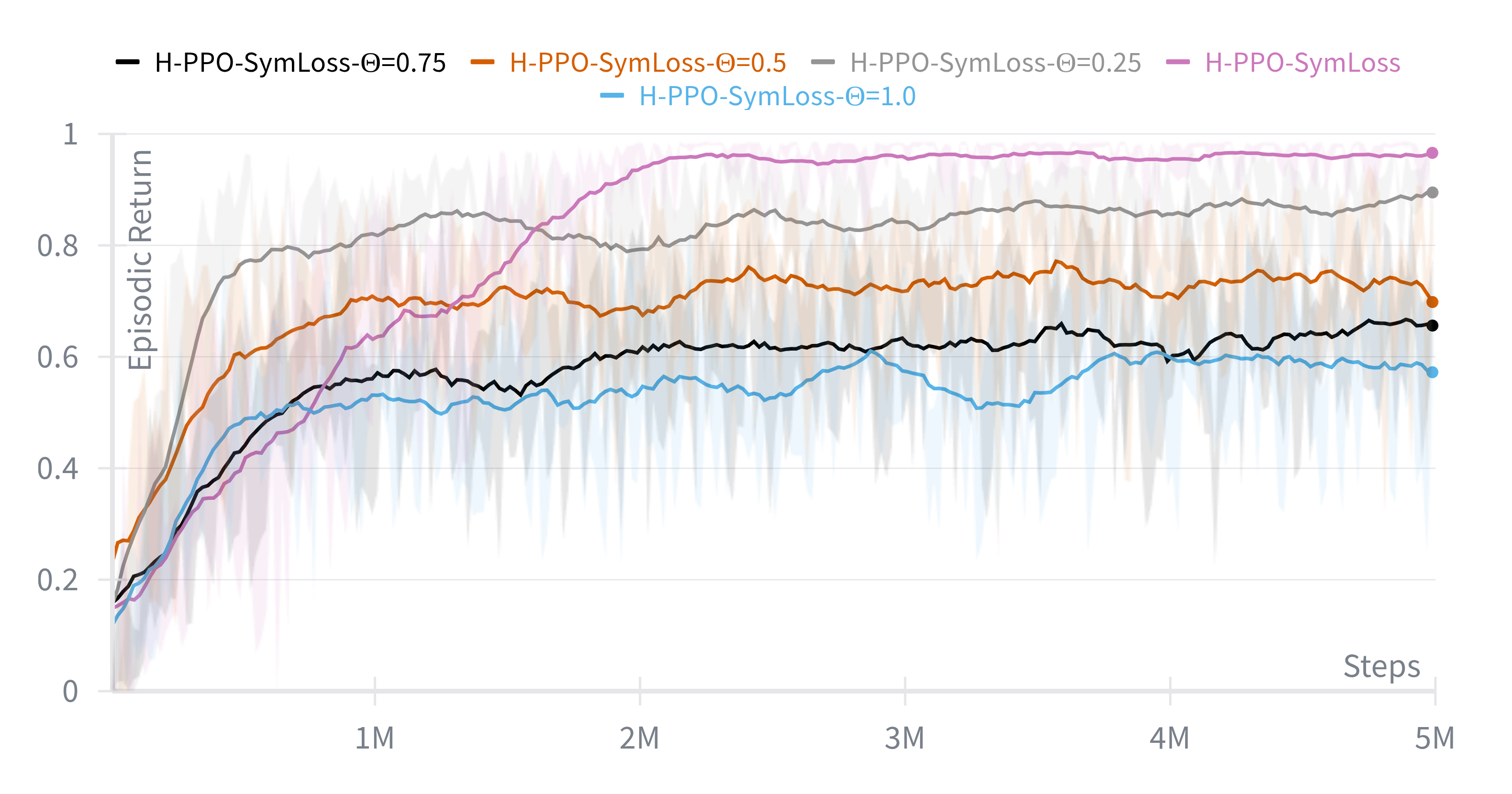}%
    }\\[1em]
    \subfigure[DeliverCoffeeAndMail]{%
        \includegraphics[width=0.49\linewidth]{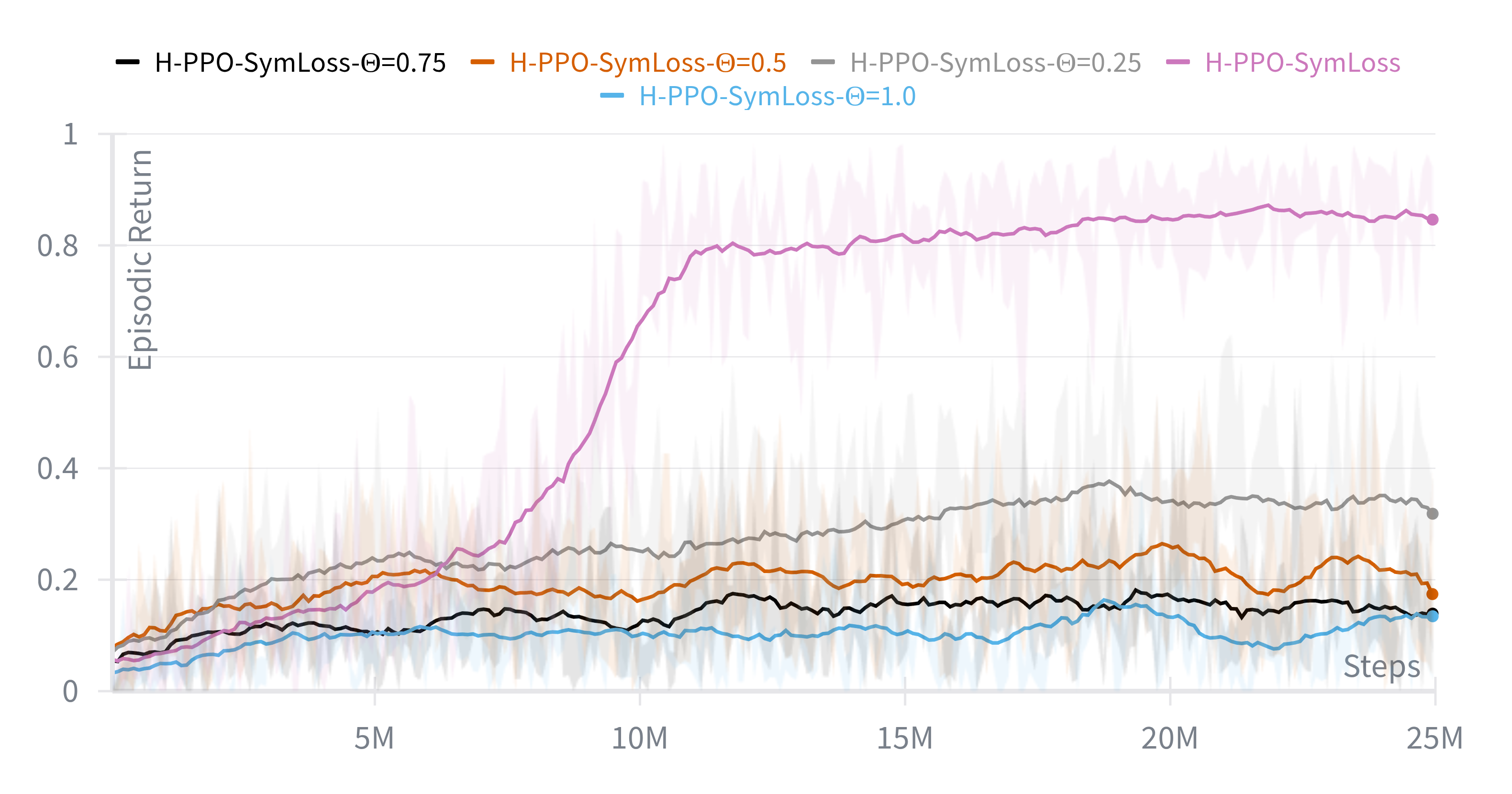}%
    }\hfill
    \subfigure[PatrolABC]{%
        \includegraphics[width=0.49\linewidth]{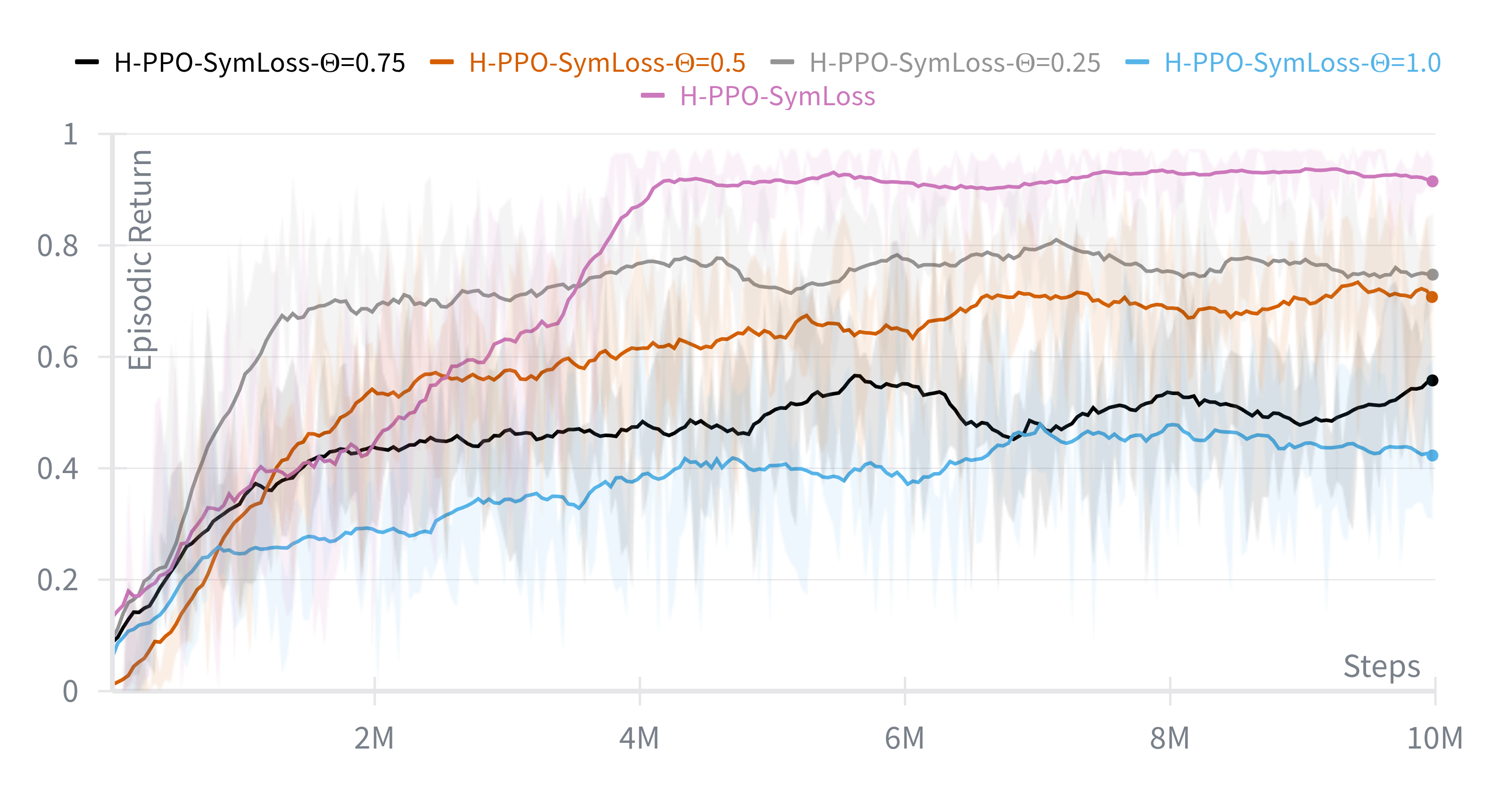}%
    }
    \caption{Ablation charts for \(\Theta\) in OfficeWorld \label{fig:ablation_theta_officeworld}}
\end{figure*}

\begin{figure*}
    \centering
    \subfigure[RedGreen]{%
        \includegraphics[width=0.32\linewidth]{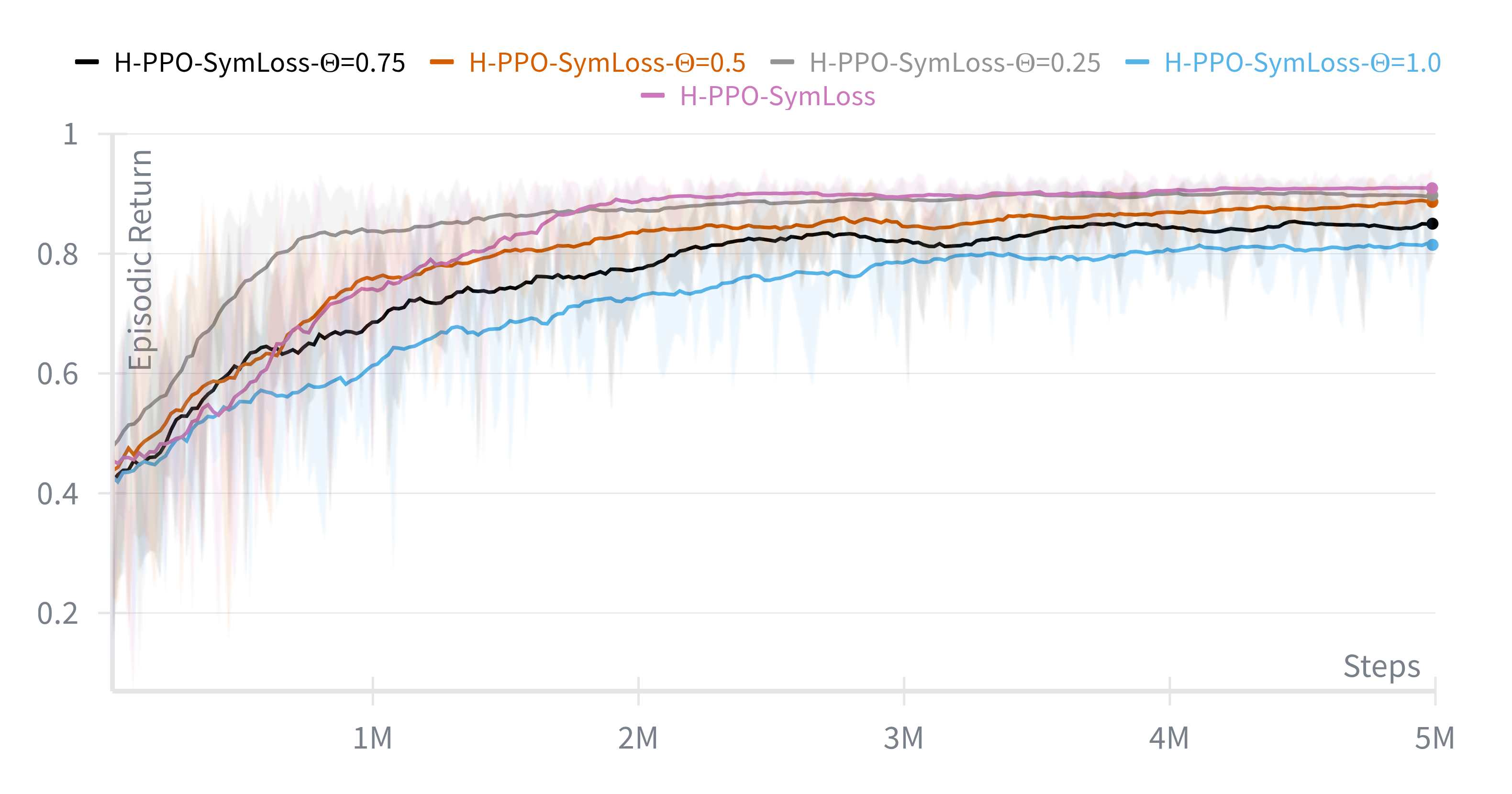}%
    }\hfill
    \subfigure[RedGreenAndBlueCyan]{%
        \includegraphics[width=0.32\linewidth]{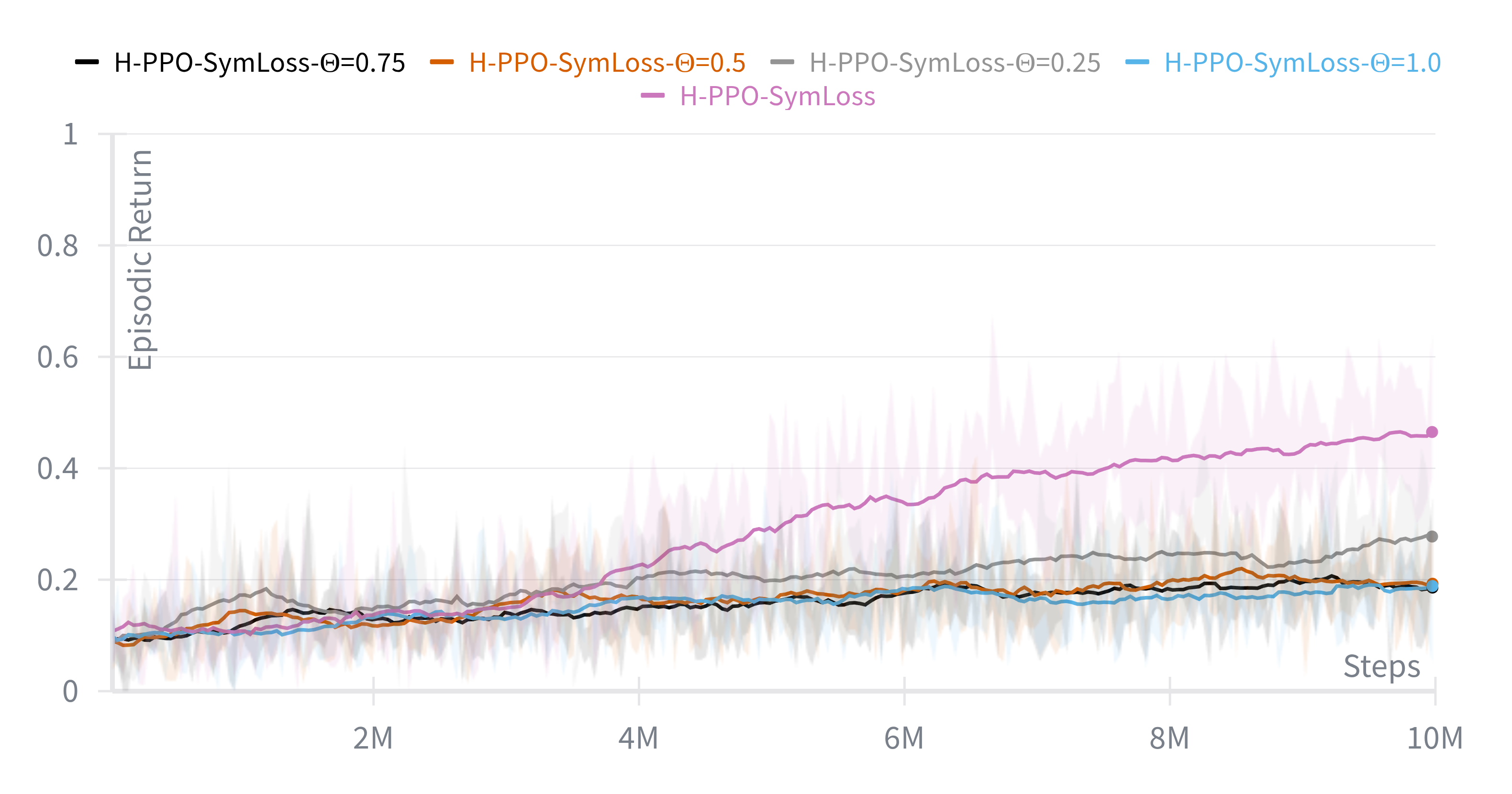}%
    }\hfill
    \subfigure[RedGreenAndBlueCyan\newline AndMagentaYellow]{%
        \includegraphics[width=0.32\linewidth]{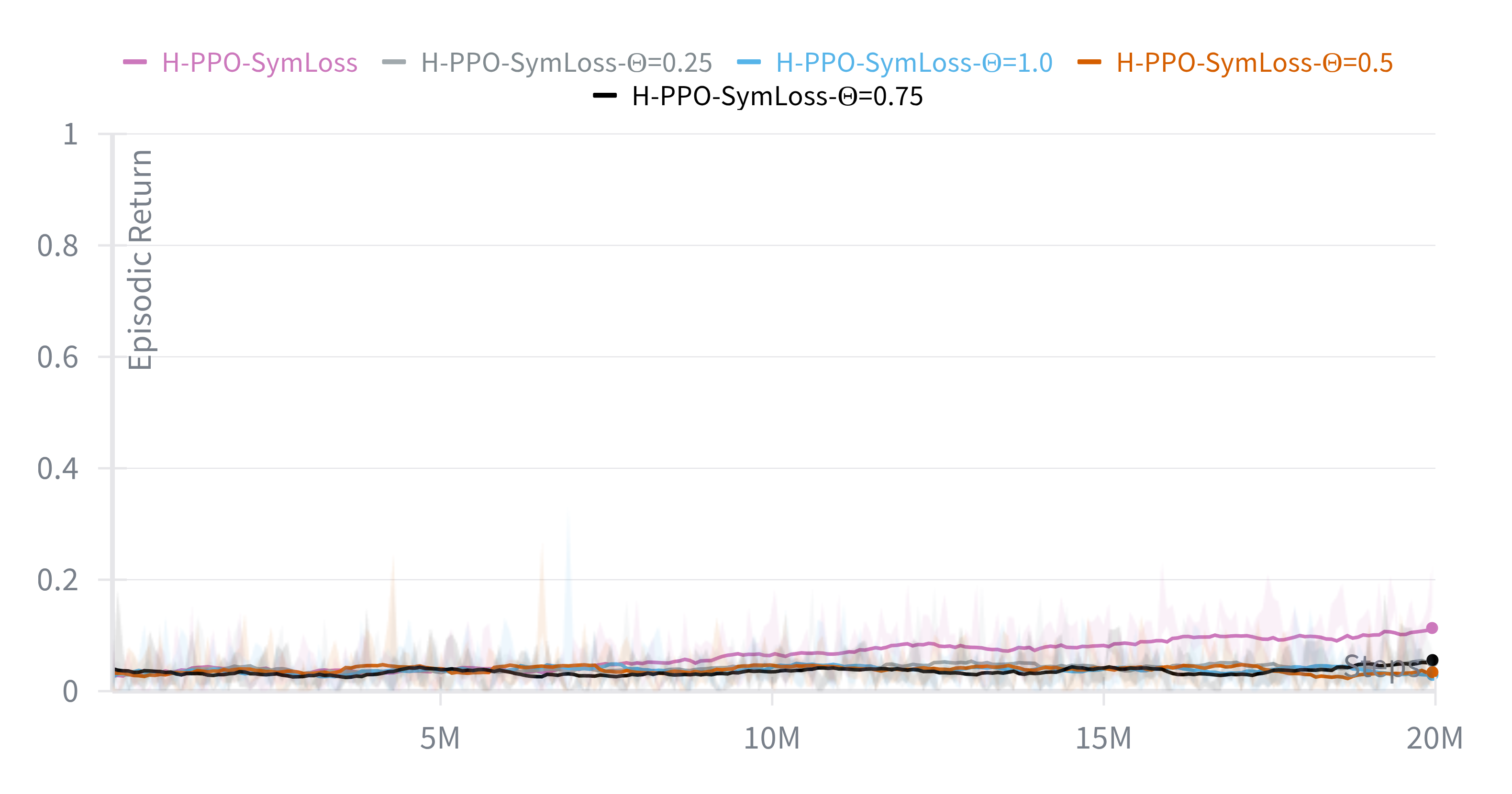}%
    }
    \caption{Ablation charts for \(\Theta\) in WaterWorld \label{fig:ablation_theta_waterworld}}
\end{figure*}

We investigated two regimes for $\Theta$ in \textsc{H-PPO-SymLoss}: 
\begin{itemize}
    \item a linearly decaying schedule \(\Theta_t = \max(\Theta_i - t \cdot \Theta_r, \Theta_f)\) with $\Theta_i=1$, $\Theta_f=0$ and $\Theta_r=0.4$  (standard \textsc{H-PPO-SymLoss} in the main paper);
    \item fixed constant weights $\Theta \in \{0.25, 0.5, 0.75, 1.0\}$.
\end{itemize} 
Across all domains, our results demonstrate that maintaining a high, constant symbolic penalty often suppresses the necessary initial exploration, leading to suboptimal convergence. On the other hand, lower constant weights allow for more exploration but offer weaker eventual regularization. The decaying $\Theta_t$ schedule consistently achieves the best balance, successfully combining early exploration with strong subsequent policy stabilization. The results are reported in Figures~\ref{fig:ablation_theta_doorkey},~\ref{fig:ablation_theta_officeworld},~\ref{fig:ablation_theta_waterworld}.

\newpage
\subsection{Ablation Study for \texorpdfstring{\(\varepsilon_f\)}{epsilon final} in \textsc{H-PPO-Product}}
\label{apd:epsilon}
\begin{figure*}[h]
    \centering
    \subfigure[DoorKey $8\times 8$ - 1 key]{%
        \includegraphics[width=0.32\linewidth]{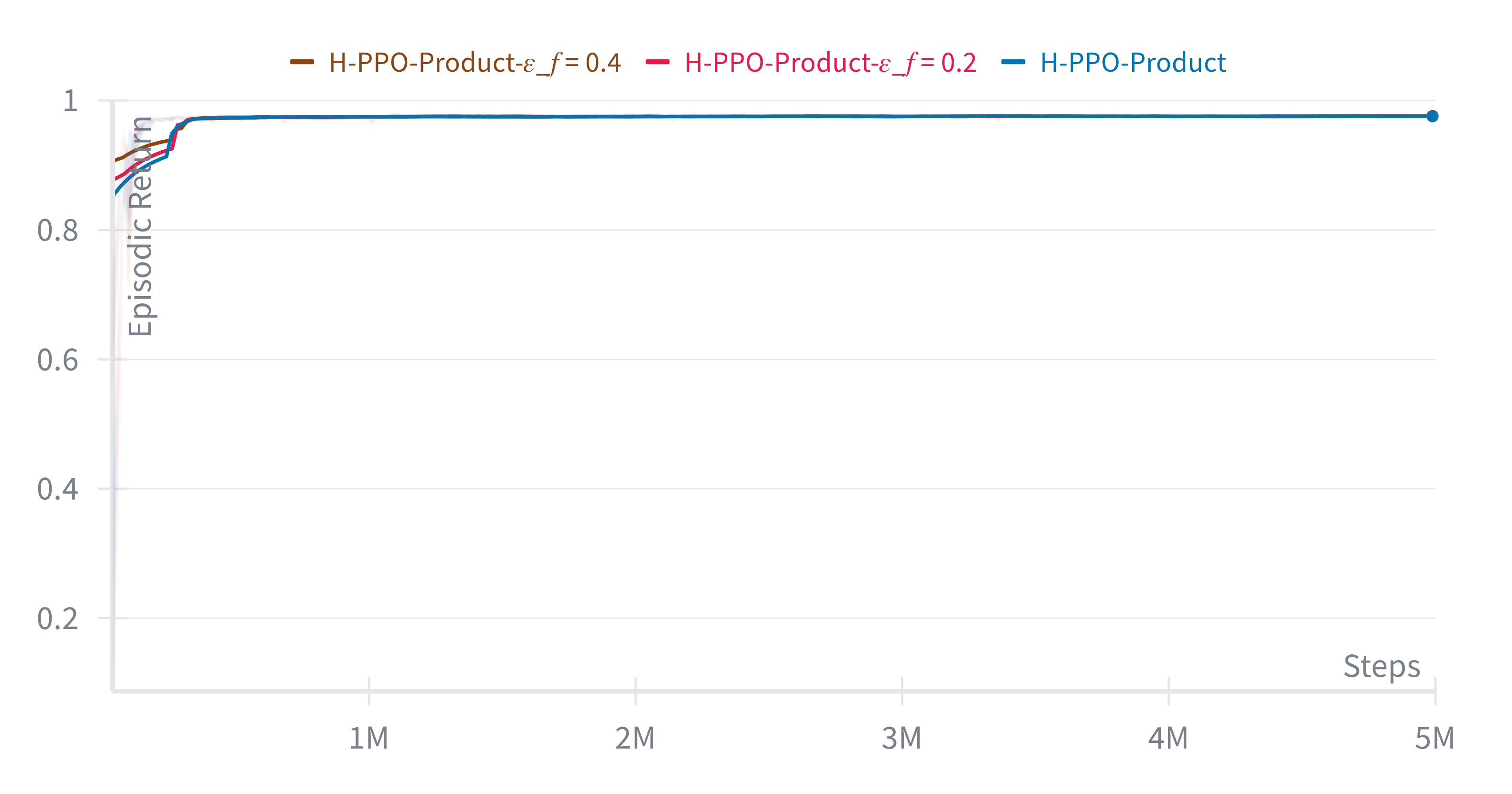}%
    }\hfill
    \subfigure[DoorKey $8\times 8$ - 2 keys]{%
        \includegraphics[width=0.32\linewidth]{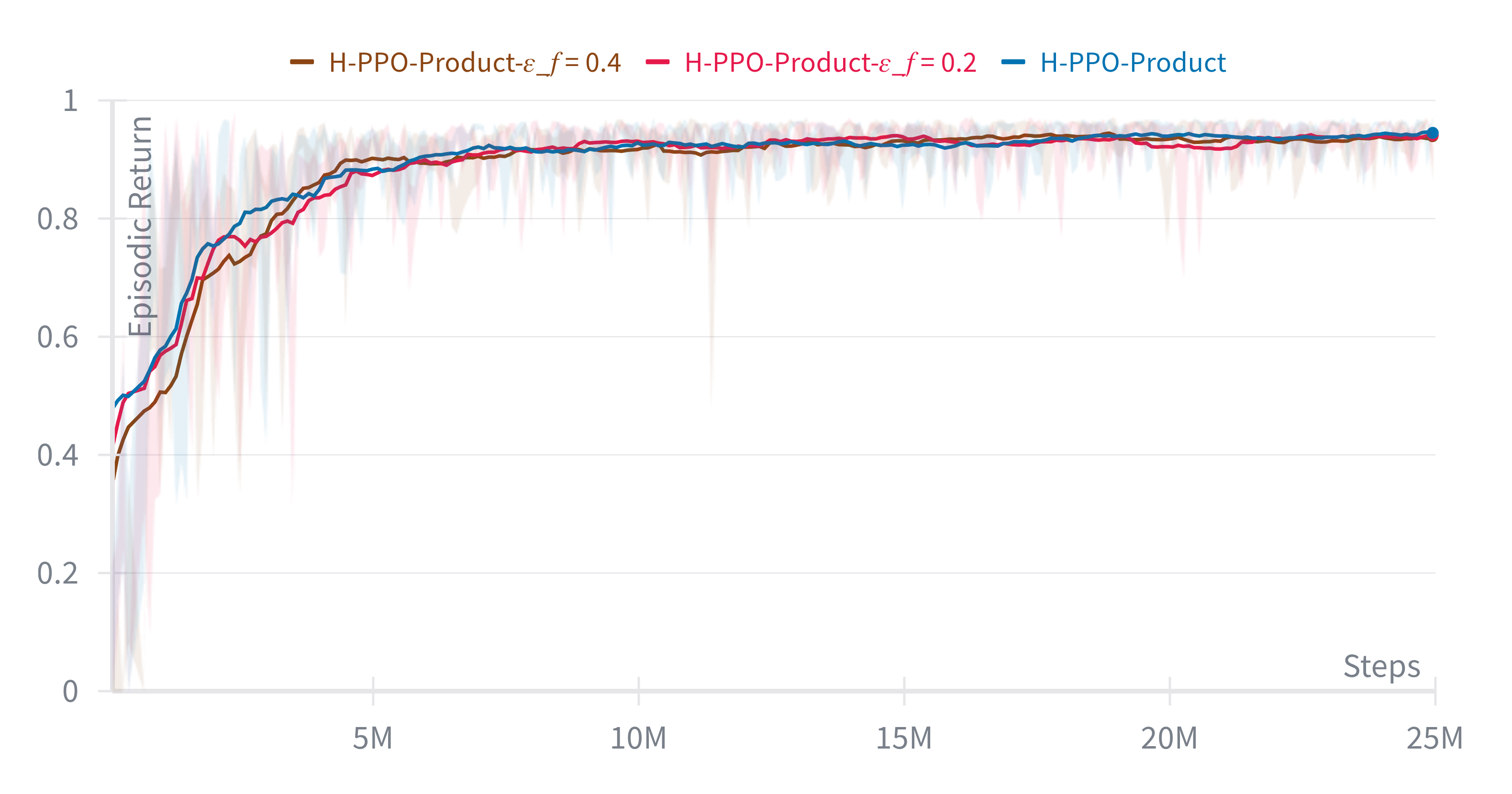}%
    }\hfill
    \subfigure[DoorKey $8\times 8$ - 4 keys]{%
        \includegraphics[width=0.32\linewidth]{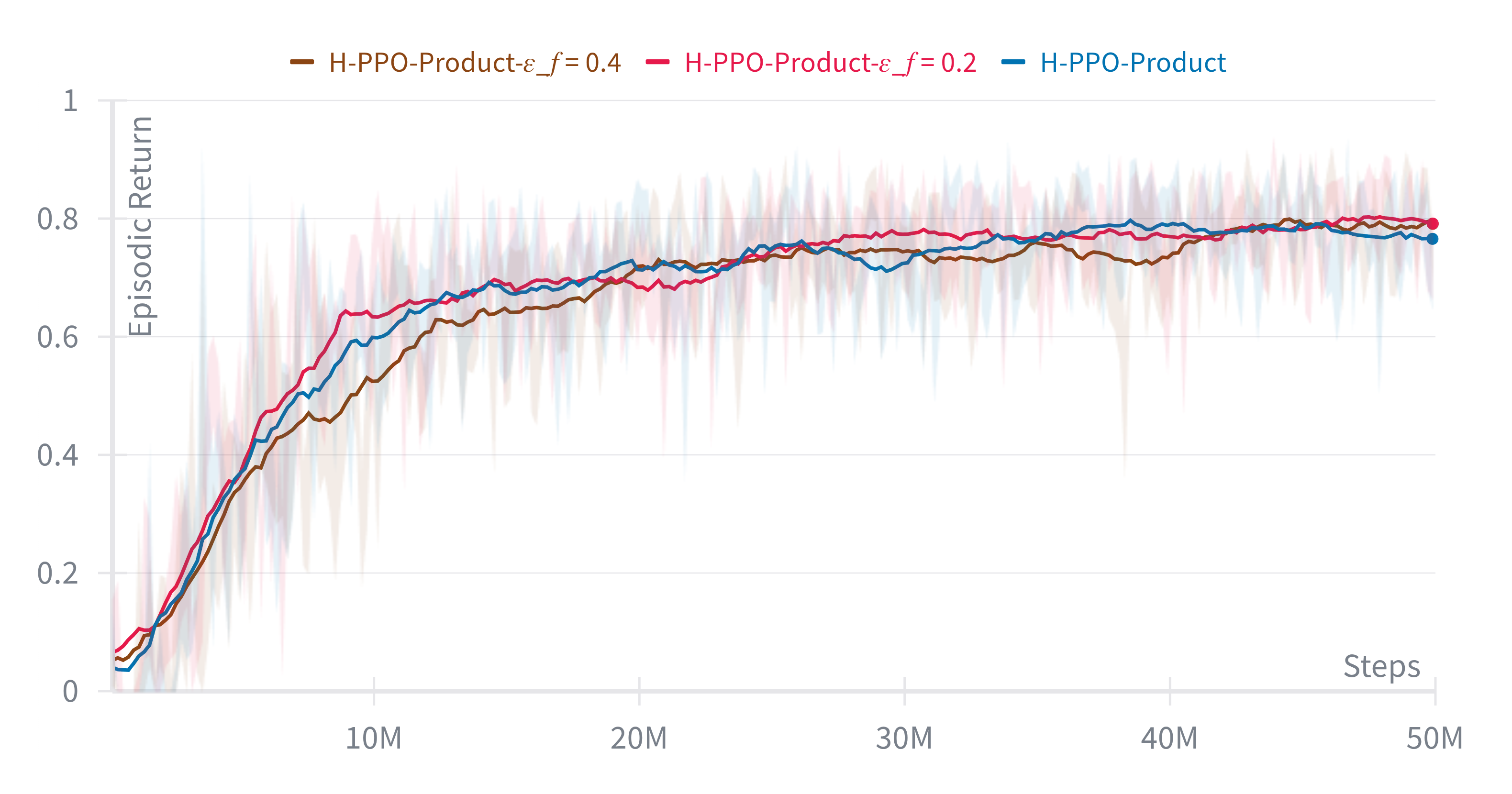}%
    }\\[1em]
    \subfigure[DoorKey $16\times 16$ - 1 key]{%
        \includegraphics[width=0.32\linewidth]{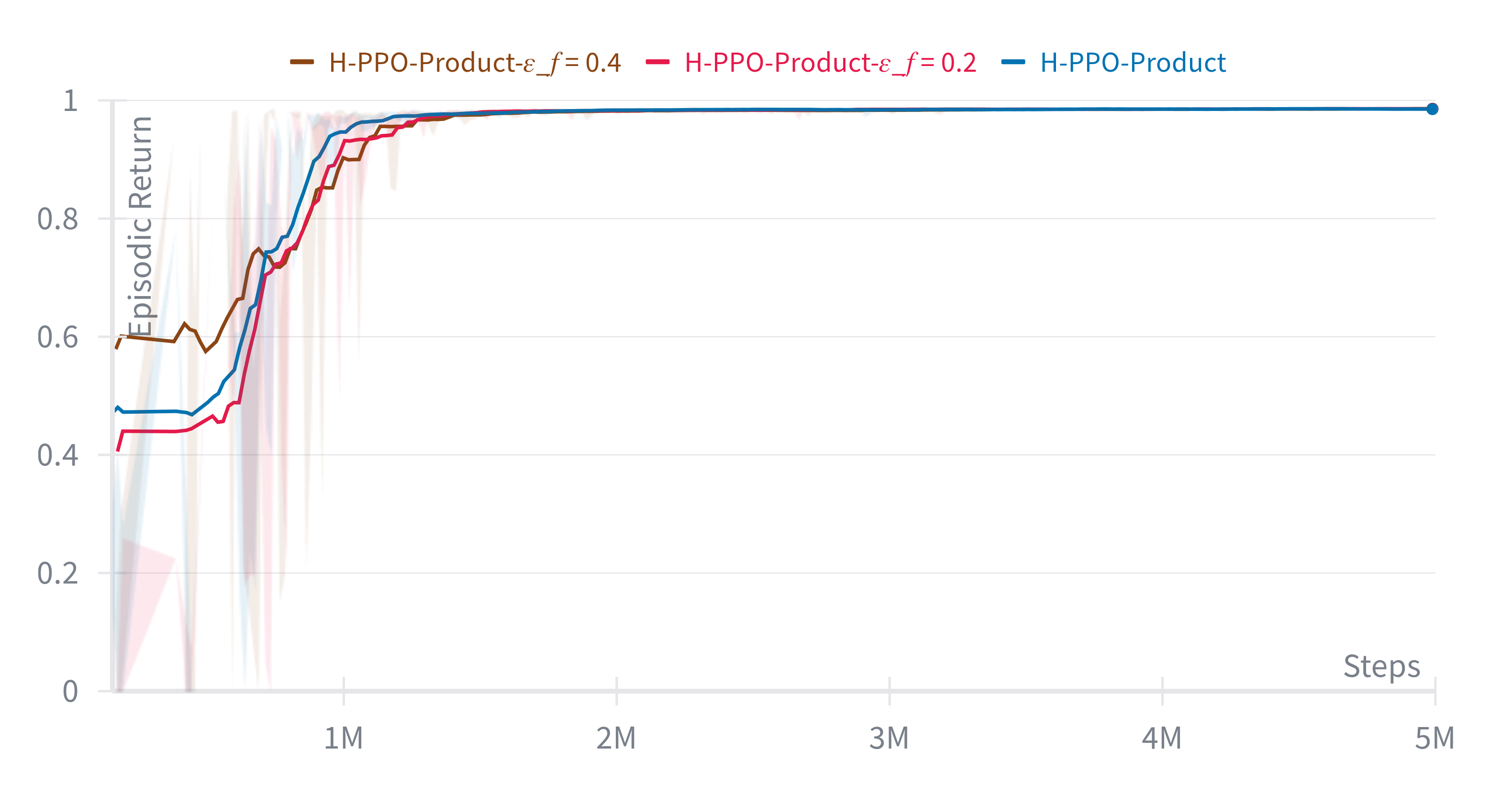}%
    }\hfill
    \subfigure[DoorKey $16\times 16$ - 2 keys]{%
        \includegraphics[width=0.32\linewidth]{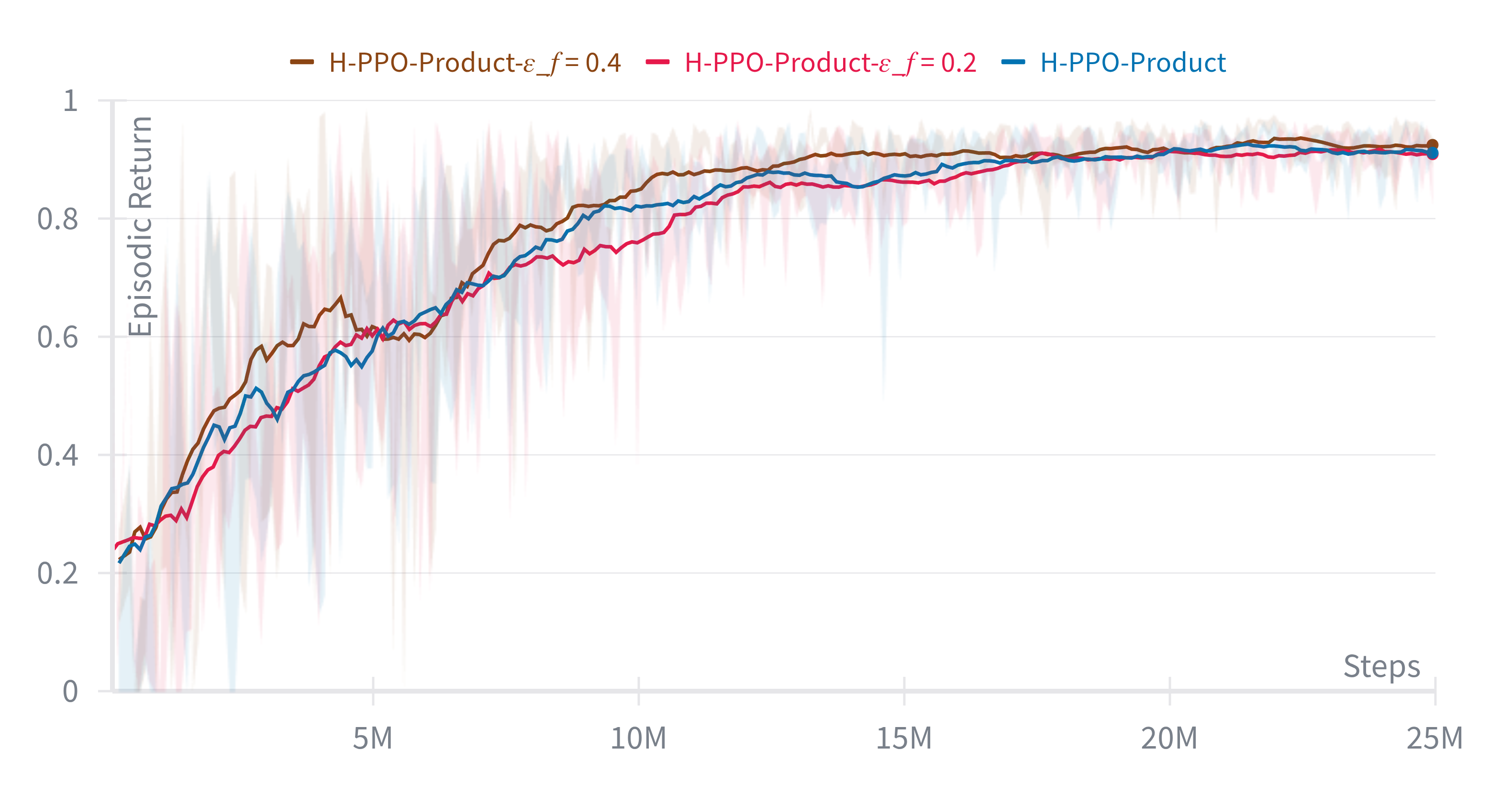}%
    }\hfill
    \subfigure[DoorKey $16\times 16$ - 4 keys]{%
        \includegraphics[width=0.32\linewidth]{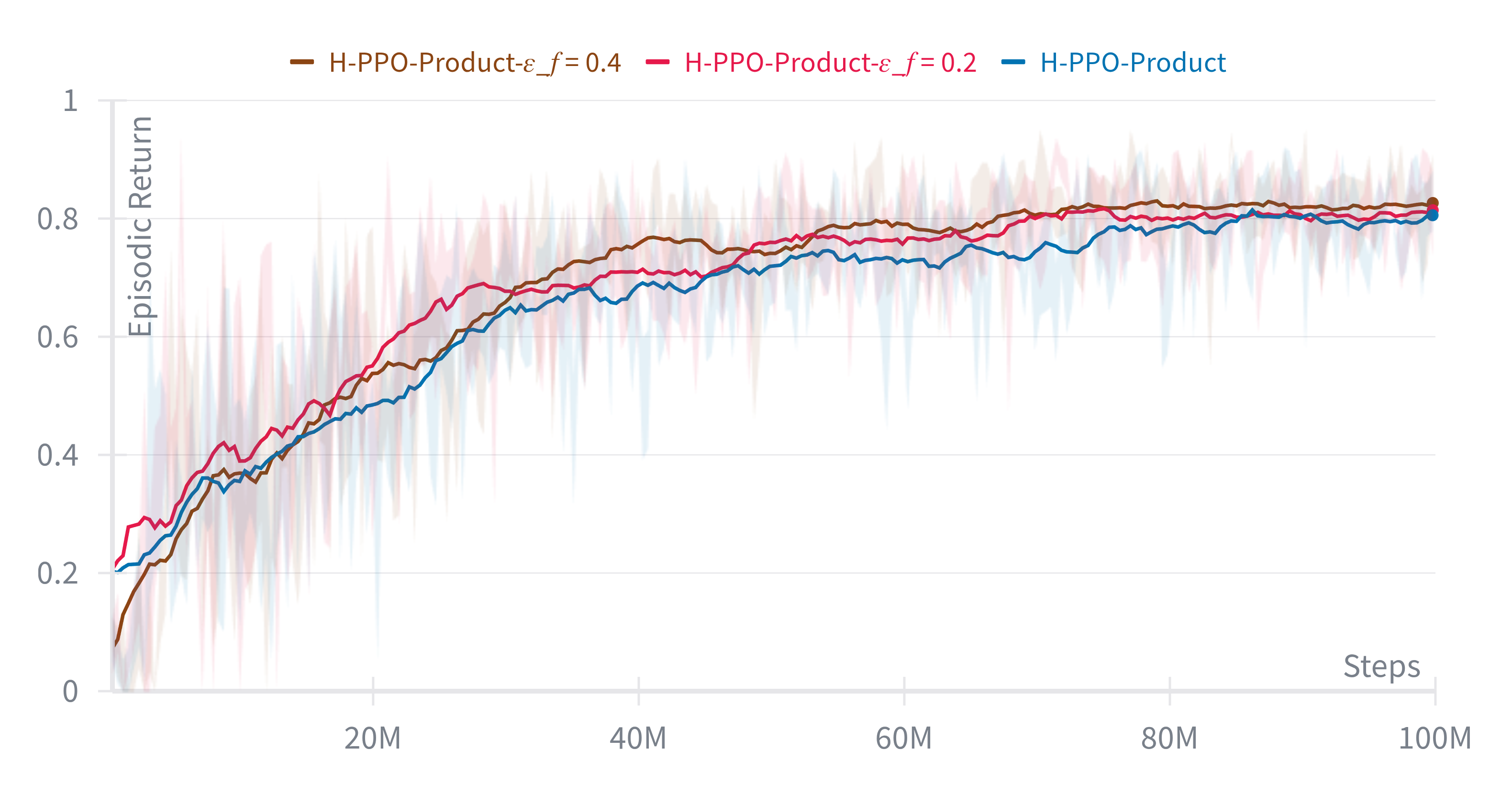}%
    }
    \caption{Ablation charts for \(\varepsilon_f\) in DoorKey \label{fig:ablation_ef_doorkey}}
\end{figure*}

\begin{figure*}
    \centering
    \subfigure[DeliverCoffee]{%
        \includegraphics[width=0.49\linewidth]{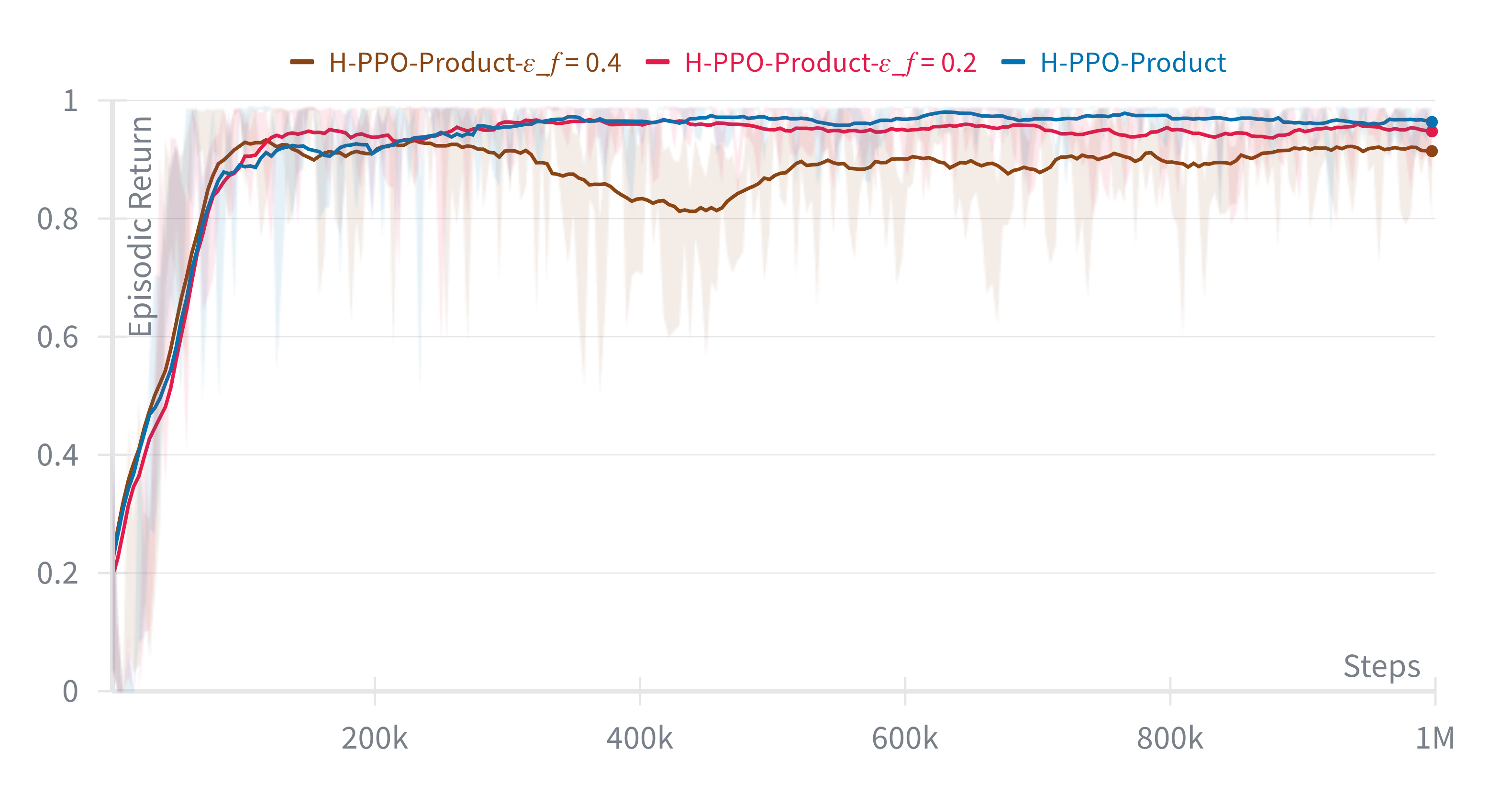}%
    }\hfill
    \subfigure[PatrolAB]{%
        \includegraphics[width=0.49\linewidth]{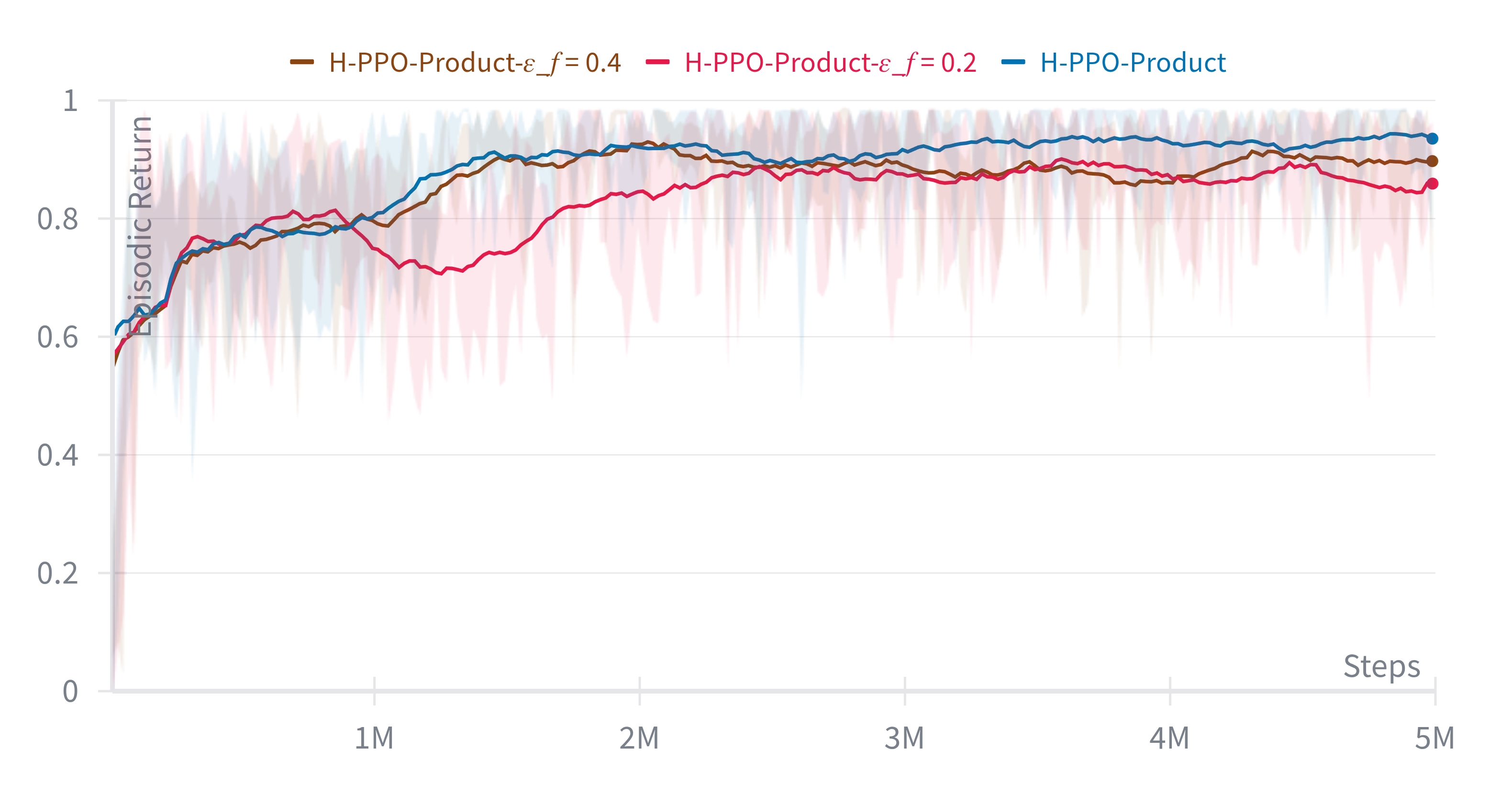}%
    }\\[1em]
    \subfigure[DeliverCoffeeAndMail]{%
        \includegraphics[width=0.49\linewidth]{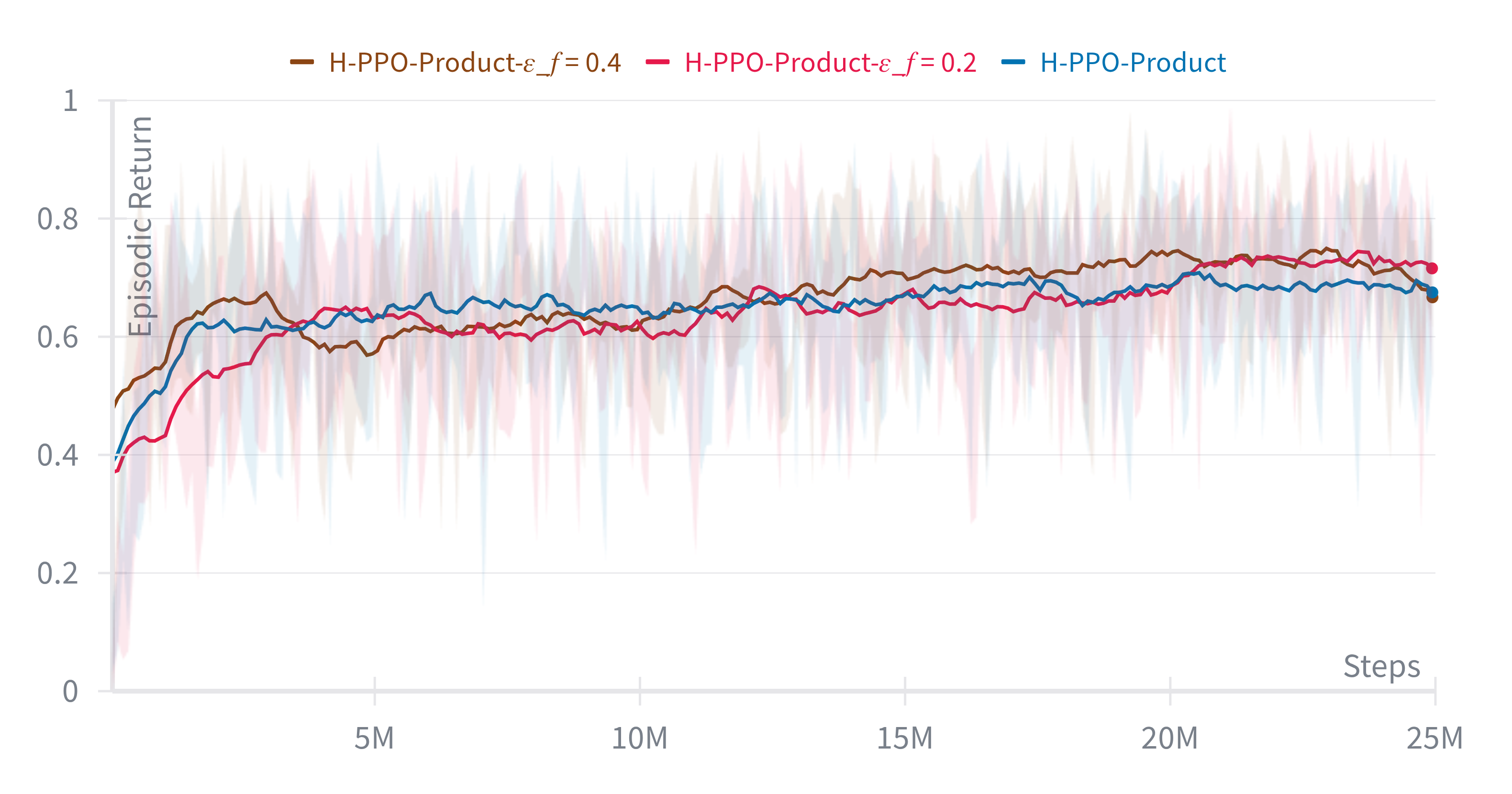}%
    }\hfill
    \subfigure[PatrolABC]{%
        \includegraphics[width=0.49\linewidth]{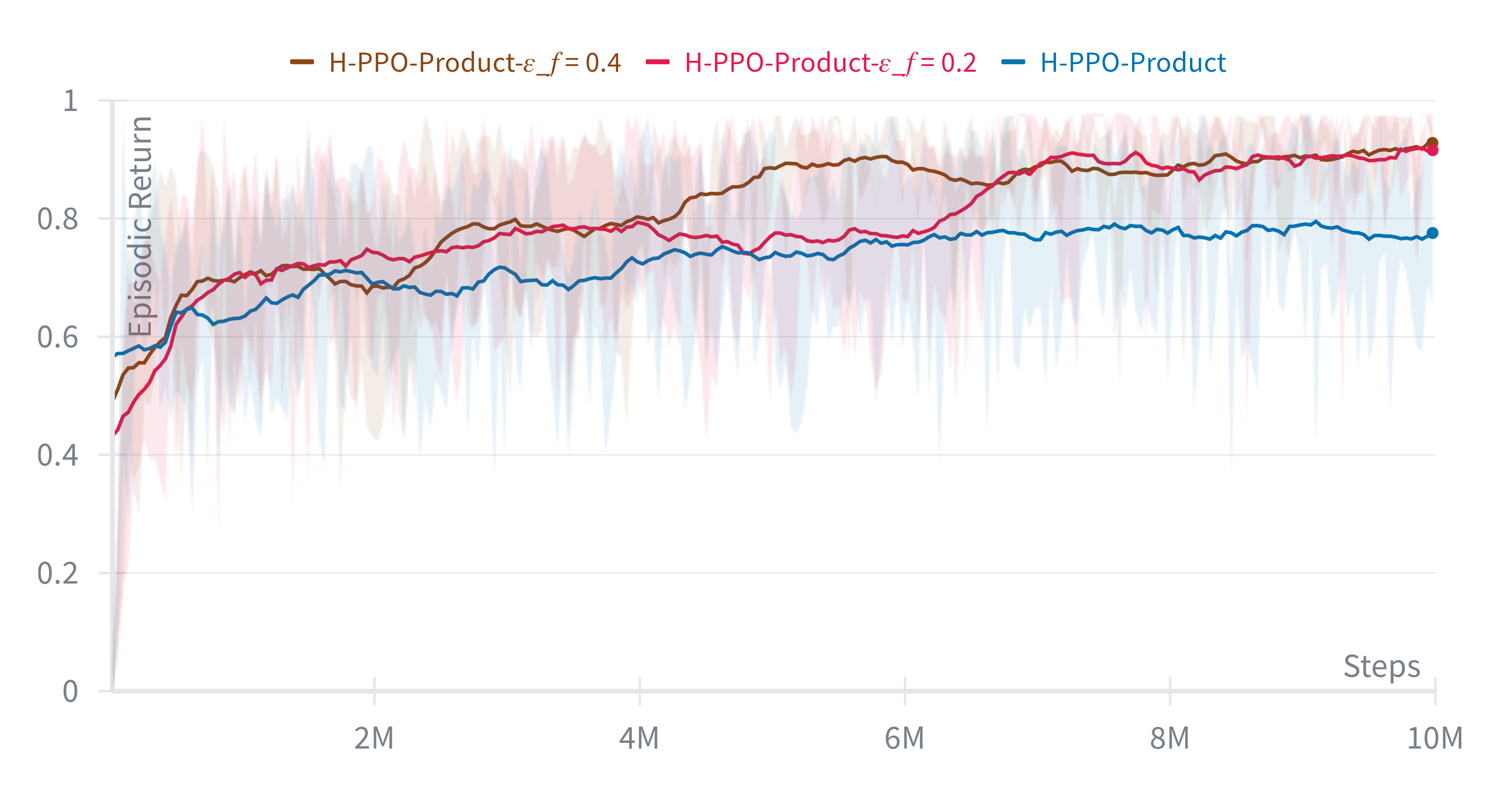}%
    }
    \caption{Ablation charts for \(\varepsilon_f\) in OfficeWorld \label{fig:ablation_ef_officeworld}}
\end{figure*}

\begin{figure*}
    \centering
    \subfigure[RedGreen]{%
        \includegraphics[width=0.32\linewidth]{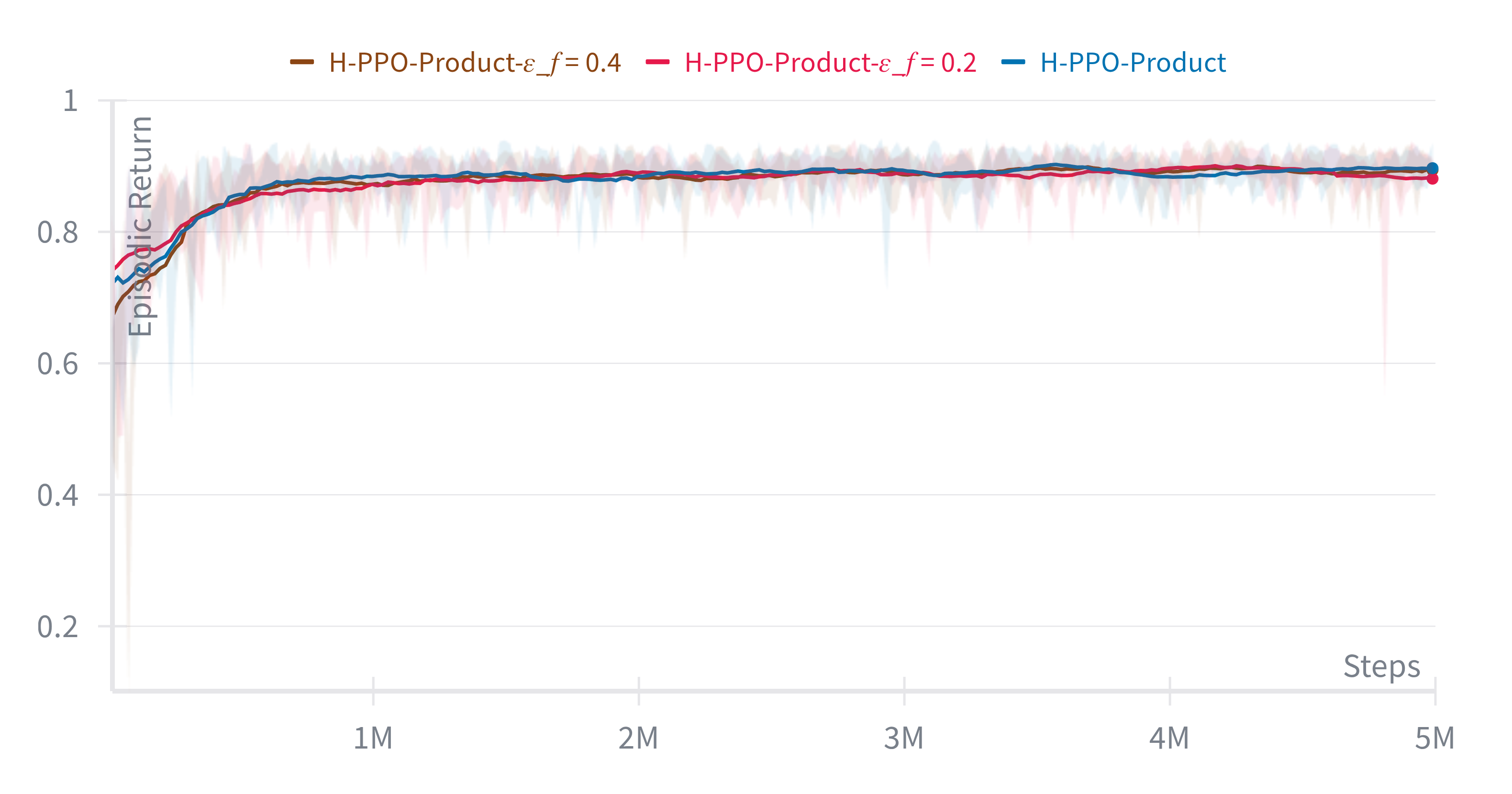}%
    }\hfill
    \subfigure[RedGreenAndBlueCyan]{%
        \includegraphics[width=0.32\linewidth]{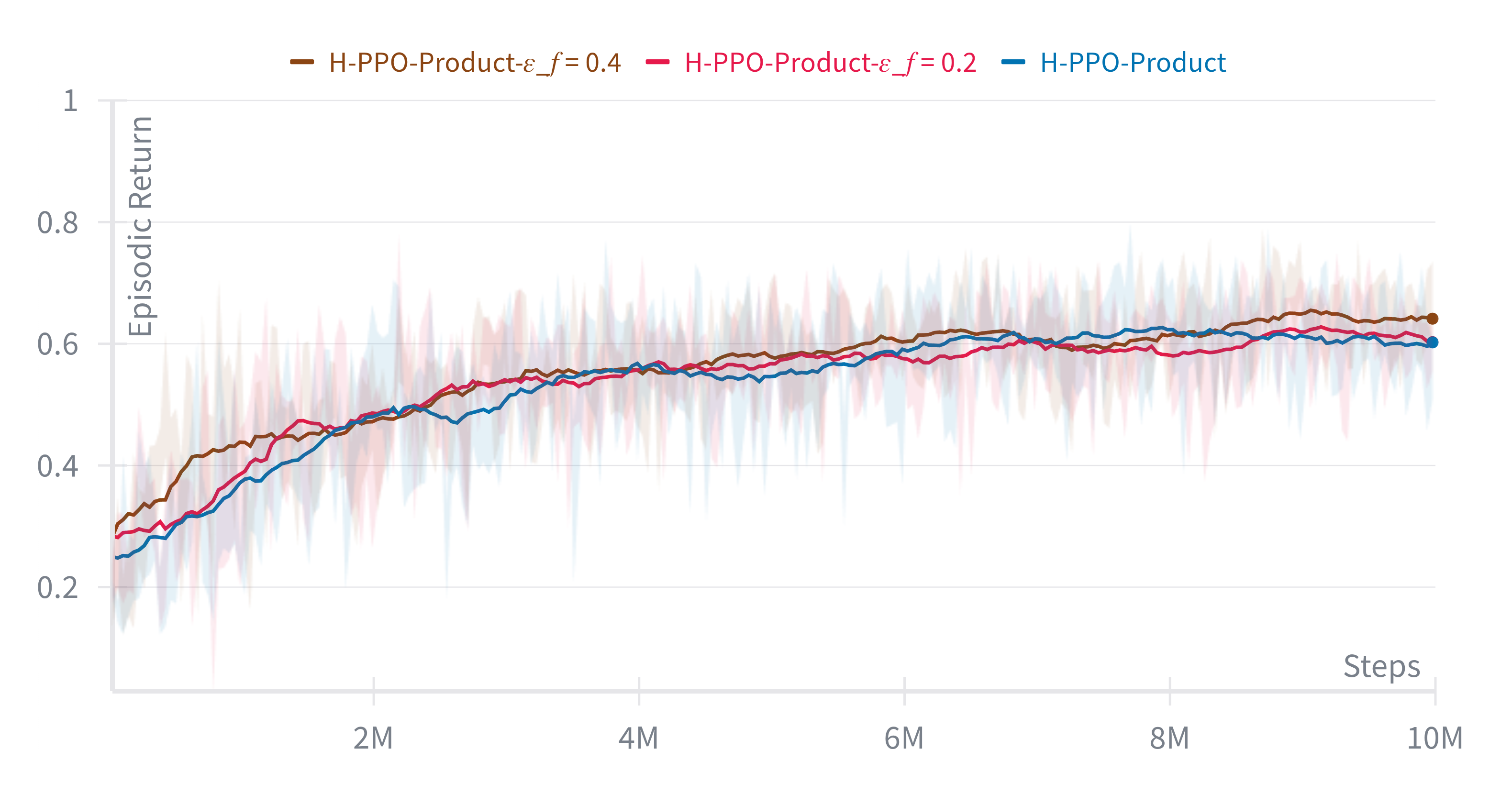}%
    }\hfill
    \subfigure[RedGreenAndBlueCyan\newline AndMagentaYellow]{%
        \includegraphics[width=0.32\linewidth]{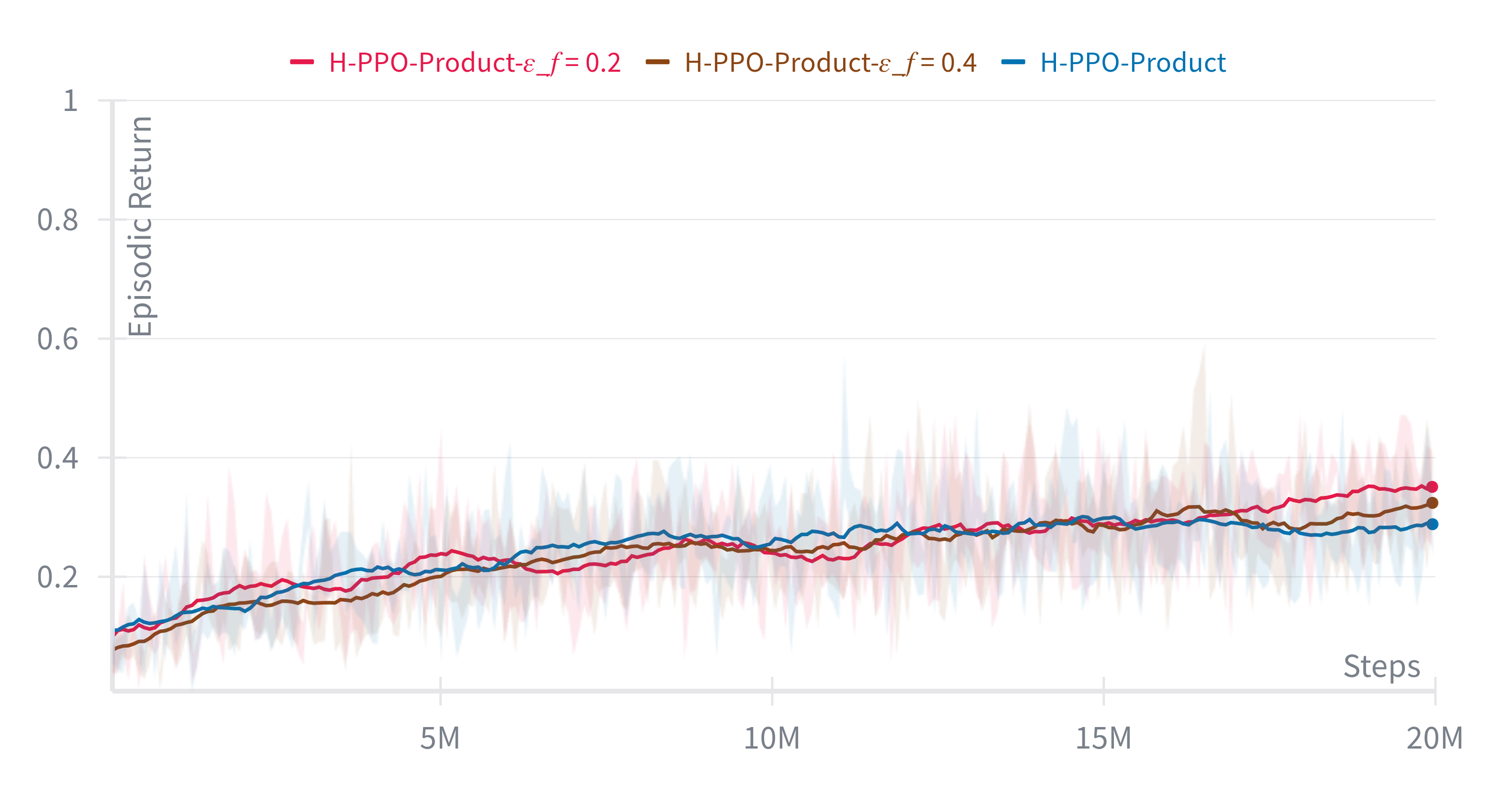}%
    }
    \caption{Ablation charts for \(\varepsilon_f\) in WaterWorld \label{fig:ablation_ef_waterworld}}
\end{figure*}

We study the effect of the symbolic influence in \textsc{H-PPO-Product} by fixing \(\varepsilon_f\) while keeping the linear decay schedule unchanged, i.e., \(\varepsilon_t = \max(\varepsilon_i - t \cdot \varepsilon_r, \varepsilon_f)\) with \(\varepsilon_i = 0\) and \(\varepsilon_r = 0.4\).
In particular, we compare three constant terminal values:
\begin{itemize}
    \item \(\varepsilon_f = 0\) (standard \textsc{H-PPO-Product} in the main paper);
    \item \(\varepsilon_f = 0.2\);
    \item \(\varepsilon_f = 0.4\).
\end{itemize}
Overall, the impact is modest on easier instances, where all curves quickly reach similar returns (e.g., DoorKey $8\times8$ with 1 key and WaterWorld RedGreen).
As the tasks become harder, maintaining a small symbolic bias is helpful: \(\varepsilon_f=0.2\) and \(\varepsilon_f=0.4\) tend to improve stability and final performance in the largest DoorKey grids, OfficeWorld \emph{PatrolABC}, and the hardest WaterWorld sequence, while leaving the learning speed mostly unchanged.
We observe a mild trade-off in the easiest OfficeWorld task (\emph{DeliverCoffee}), where larger \(\varepsilon_f\) can slightly reduce the final return, suggesting that a nonzero \(\varepsilon_f\) is most beneficial in long-horizon, sparse-reward regimes.
Given that these differences are limited in most tasks, we keep \(\varepsilon_f=0\) in Section~\ref{sec:experiments}. The results are reported in Figures~\ref{fig:ablation_ef_doorkey},~\ref{fig:ablation_ef_officeworld},~\ref{fig:ablation_ef_waterworld}.

\end{document}